\documentclass[letterpaper]{article}

\usepackage{times}
\usepackage{microtype}
\usepackage{amsmath,amssymb}
\usepackage{algorithm}
\usepackage{algorithmic}
\usepackage{booktabs}
\usepackage{graphicx}
\usepackage{array}
\usepackage{multirow}
\usepackage[hidelinks]{hyperref}
\usepackage[numbers,sort&compress]{natbib}
\usepackage{xcolor}

\newcommand{\thigh}{\tau}                 % single gate threshold on the margin
\newcommand{\tlow}{\tau_{\mathrm{lo}}}    % optional early-exit knob
\newcommand{\cm}{m}                        % coverage margin symbol

\title{CoVeR: Coverage-Based Routing of Verifier Calls in Agentic Retrieval}
\author{
  Daeyoung Roh \\
  Independent Researcher \\
  \texttt{dybroh@gmail.com}
  \and
  Donghee Han \\
  KAIST \\
  \texttt{venzino.han@gmail.com}
}
\date{}

\begin{document}
\maketitle

% ======================================================================
\begin{abstract}
An agentic retrieval system issues a sequence of search queries and must decide,
at each step, whether the evidence collected so far is enough to stop.
Delegating that decision to an LLM verifier or a prompt judge makes stopping
reliable, but the verifier then reprocesses the growing evidence after every
retrieval step, a substantial repeated cost. We show that most of these calls
can be skipped without materially changing answer accuracy: a single threshold on a
frozen sentence-embedding \emph{coverage margin} detects the states in which the
evidence is still plainly incomplete, and the verifier is called only on the
ambiguous remainder --- a gate we call CoVeR (\textbf{Co}verage-based
\textbf{Ve}rifier \textbf{R}outing). Across three multi-hop QA benchmarks, with the evaluation
protocol fixed before the full-scale run, the CoVeR-gated agent matches the
answer accuracy of both the full-budget agent and the always-verify baseline
within a fraction of an EM point. It cuts $62$--$68\%$ of
verifier calls, and $93\%$ in a
saturated regime. Routers built on evidence counts, lexical overlap, or BM25
relevance, alone or learned in combination, give weaker overall trade-offs, the gate transfers without re-tuning
across deciders and agent scales, and its drafter distills into a $921$k-parameter
head atop the frozen encoder, leaving no LLM in the routing loop. The same signal cannot \emph{replace}
verification: matching a claim is far easier than deciding the claim is
supported.
\end{abstract}

% ======================================================================
\section{Introduction}
\label{sec:intro}

A retrieval-augmented search agent answers a multi-hop question by issuing a
sequence of queries, each retrieving new evidence, until it decides it has
enough to answer \citep{lewis2020rag,press2023selfask,yao2023react,jin2025searchr1,chen2025research}.
When to stop is a control problem: stopping early can miss a required reasoning
hop, while continuing past sufficiency wastes retrieval calls, adds latency,
and buries the answer in distracting context. A reliable way to make the
decision is to frame stopping as \emph{evidence coverage} and delegate it to a
\emph{decider} that reads the accumulated evidence and rules on whether each
required claim is now supported, either with a trained coverage verifier or
with a zero-shot sufficiency judge
\citep{yang2025simrag,park2025stop,asai2024selfrag}. The catch is that the
decider is itself an LLM, and it is consulted after every retrieval step. Once
the stopping policy is fixed, repeatedly invoking the decider becomes an
avoidable verification overhead: in our setting, nine to eleven claim
verdicts per question, each a full read of the accumulated evidence
(\S\ref{sec:results-eff}).

\begin{figure}[t]
\centering
\includegraphics[width=0.55\linewidth]{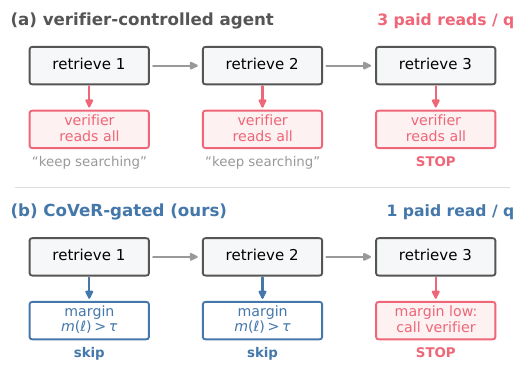}
\caption{\textbf{Routing the verification bill.} (a)~The verifier reads all
evidence at every step. (b)~CoVeR pays only for the ambiguous step (outcome
preservation: \S\ref{sec:results-route}). Source: \texttt{results/raw/e2/}.}
\label{fig:contrast}
\end{figure}

Much of that reading is spent on states whose verdict is not in doubt.
Consider a 2WikiMultihopQA question: \emph{``Which film has the director died
earlier, Remember the Day or Cast Up by the Sea?''} The agent needs each
film's director and death date. Its first two steps retrieve \emph{Remember
the Day}'s side (Henry King, d.~1982), but nothing yet links \emph{Cast Up by
the Sea} to any director, and a verifier-controlled run pays for two ``keep
searching'' verdicts (Figure~\ref{fig:contrast}). Both were predictable
without an LLM read: the least-covered claim's margin holds at $0.18$ against
a threshold of $0.16$, so CoVeR skips both calls. At the third step the
missing link arrives (\emph{``Cast Up by the Sea is a 1916 Australian film
directed by John Gavin''}, d.~1938), the margin drops to $0.13$, the gate pays
for one verdict, the verifier fires --- and every arm returns the correct
answer (the full trajectory, with qid, is in Appendix~\ref{app:case}).

Our method operationalizes this observation. Embed each expected hop claim and each
retrieved sentence with a frozen encoder, and define a per-claim
\emph{coverage margin} as one minus the best cosine similarity to any evidence
retrieved so far; the state margin is the largest per-claim margin. A single
threshold $\thigh$ routes the decider: while the least-covered claim is still
clearly uncovered, the verdict is obviously ``keep searching'' and we skip the
call; otherwise we pay for a verdict and follow it. A skip cannot introduce an
unverified stop; it can only defer stopping to a later, adjudicated loop.
Deferral can still change what the agent reads and hence its answer, an effect
our evaluation measures directly (\S\ref{sec:results-route}), but every stop
still passes through the LLM decider.

We organize the paper around two questions. Can the margin simply
\emph{replace} the decider? No. With oracle evidence targets it separates
covered from uncovered claims well, but the deployable version built on
generated claim strings falls far short of that ceiling, a label-free predictor
trained to close the gap fails, and a margin-only stopping rule forced to
match the verifier's safety degenerates to never stopping
(\S\ref{sec:results-replace}). Deciding that evidence supports a claim requires
reading how entities and relations compose; cosine proximity does not establish
it. This is the matching-is-not-supporting gap familiar from natural-language
inference and attribution evaluation
\citep{thorne2018fever,bohnet2022autoais,gao2023alce}, measured here for
coverage stopping.

Can the margin \emph{route} the decider? Yes, and this is the main result. With
thresholds selected leave-one-dataset-out, the CoVeR-gated agent matches the
accuracy of the full-budget agent and the always-verify baseline --- every
$95\%$ CI on the EM difference lies within $\pm1.2$\,pp --- at $62$--$68\%$ fewer verifier calls
(\S\ref{sec:results-route}). The saving is not available to cheaper signals:
routers built on evidence counts, lexical overlap, or BM25 relevance cannot
reach the same call cut at the same safety bar (\S\ref{sec:results-ctl}). Nor is
it tied to one decider or one scale: the same threshold, with no re-tuning,
routes a public prompt judge and a 7B agent and passes an initial open-corpus
check, and the gate-side claim embeddings distill into a sub-$1$M drafter
that leaves no LLM in the routing path (\S\ref{sec:results-transfer}).

We further analyze whether the routing signal can also veto erroneous
decider stops; the data supports this only directionally. The margin vetoes a volatile prompt judge's premature stops,
a consistent but not statistically significant gain; it cannot audit the trained
verifier, whose false stops are caused by the same deceptive similarity that
fools the margin itself (\S\ref{sec:analysis}).

\paragraph{Contributions.}
\begin{itemize}
\item \textbf{One-sided verifier routing.} A regime that skips only clearly
futile verifier calls while every stopping decision remains decider-issued
--- viable because matching a claim is easier than judging it supported
(\S\ref{sec:results-replace}).
\item \textbf{CoVeR.} An instantiation by a single untuned threshold on a
frozen embedding coverage margin: it cuts $62$--$68\%$ of verifier calls at
EM within a fraction of a point of both references, and no cheaper or
learned router we test matches this trade-off
(\S\ref{sec:results-route}--\S\ref{sec:results-ctl}).
\item \textbf{Transfer and limits.} The same gate routes three deciders and
a larger agent without re-tuning, and distills into a $921$k-parameter
drafter whose claim-free pairing removes claim generation entirely
(\S\ref{sec:results-transfer}); but the margin cannot replace verification,
and it cannot correct errors it shares with the verifier
(\S\ref{sec:analysis}).
\end{itemize}

% ======================================================================
\section{Background and Related Work}
\label{sec:related}

\paragraph{Learning to defer, cascades, and confidence-gated routing.}
Our gate is a cascade invoking an expensive expert only when a cheap stage is
uncertain \citep{chen2023frugalgpt}. Unlike model cascades, whose stages differ
in size, ours differ in \emph{kind} (a geometric check versus a reading
judgment), and it repeats at every retrieval step. From learning to defer
we take one licensing point: the rejector need not be accurate on the
decision itself, only calibrated about when it is confident
\citep{madras2018predict,mozannar2020consistent} --- which is why a signal
too blunt to decide stopping (\S\ref{sec:results-replace}) can still route
it. Similar asymmetric gates appear in SAGE, which defers only
ambiguous memory writes to an LLM \citep{wang2026sage}. Closest is selective
verification for reasoning \citep{qu2026adaptive}, which allocates verifier
calls among candidate reasoning states; question-level routers pick a
retrieval strategy per question \citep{li2026raser}, and RL-finetuned agents
prune retrieval depth by question difficulty \citep{java2025frugalrag}. CoVeR
routes neither retrieval nor depth: it gates \emph{when a fixed stopping
decider is consulted}, one-sidedly, never issuing an unverified stop.

\paragraph{Matching is not supporting.}
Whether retrieved text \emph{supports} a claim is the object of NLI and
attribution evaluation: FEVER casts verification as entailment
\citep{thorne2018fever}, and AutoAIS and ALCE score whether a statement is
entailed by its cited evidence \citep{bohnet2022autoais,gao2023alce};
\citet{joren2025sufficient} lift this to \emph{sufficiency}. Our
replacement negative measures the same gap in embedding space for per-hop coverage stopping: cosine proximity is a good difficulty signal but a poor sufficiency decision.

\paragraph{Speculative execution.}
Speculative decoding drafts tokens with a small model for a large one to verify
in parallel, exactly preserving the output
\citep{leviathan2023speculative,chen2023speculative}; Speculative RAG drafts
\emph{answers} for a larger verifier \citep{wang2024speculativerag}. Speculative
decoding can check every draft because verification is one parallel forward
pass; a skipped verifier call has no parallel check, so our guarantee is
statistical rather than exact.

\paragraph{Adaptive stopping and retrieval.}
Prior work learns or defines the retrieval control action: triggering on
decoding-time uncertainty \citep{jiang2023active} or question complexity
\citep{jeong2024adaptiverag}, and, closer to stopping, ending on an answer
span \citep{trivedi2023interleaving}, a trained sufficiency critic
\citep{yang2025simrag}, or a value-based stop policy \citep{park2025stop}.
All decide the control \emph{action}; we take the policy as given and lower
the cost of delivering it, composably with any of them. To our knowledge, no
prior work routes verifier calls with a one-sided frozen coverage signal
while keeping every stop decider-issued.

% ======================================================================
\section{Method}
\label{sec:method}

\paragraph{Setup.}
A search agent runs loops $\ell=1,\dots,L$ ($L{=}6$ in our data), accumulating
evidence $S_\ell$. The question induces a small set of expected hop
\emph{claims} $\{c_1,\dots,c_k\}$ ($k\approx2$). A \emph{decider} $D$ (a trained
verifier or a prompt judge) reads $S_\ell$ each loop and rules on whether every
claim is covered; the agent stops at the first loop $D$ rules the state covered
(we say $D$ \emph{fires}). $D$ is invoked once per loop until it fires, the
cost we route.

\paragraph{Frozen-embedding coverage margin.}
Let $\phi$ be a frozen E5 sentence encoder \citep{wang2022e5}. For claim $c_j$
with target text $t_j$ and evidence $S_\ell$, define the per-claim coverage
margin and state margin
\[
  \cm_j(\ell)=1-\max_{s\in S_\ell}\cos\!\big(\phi(t_j),\phi(s)\big),\qquad
  \cm(\ell)=\max_j \cm_j(\ell).
\]
A high margin means ``no retrieved sentence is close to this claim.'' We call this score a coverage margin: a frozen similarity shortfall, not a
learned energy. The target $t_j$ is a \emph{generated} expected-claim
string (deployable, no gold labels), a gold supporting sentence (an oracle
diagnostic only), or a learned predictor's output (below). Computing $\cm(\ell)$
is a handful of cached dot products, negligible against a decider call.

\begin{figure}[t]
\centering
\includegraphics[width=0.9\linewidth]{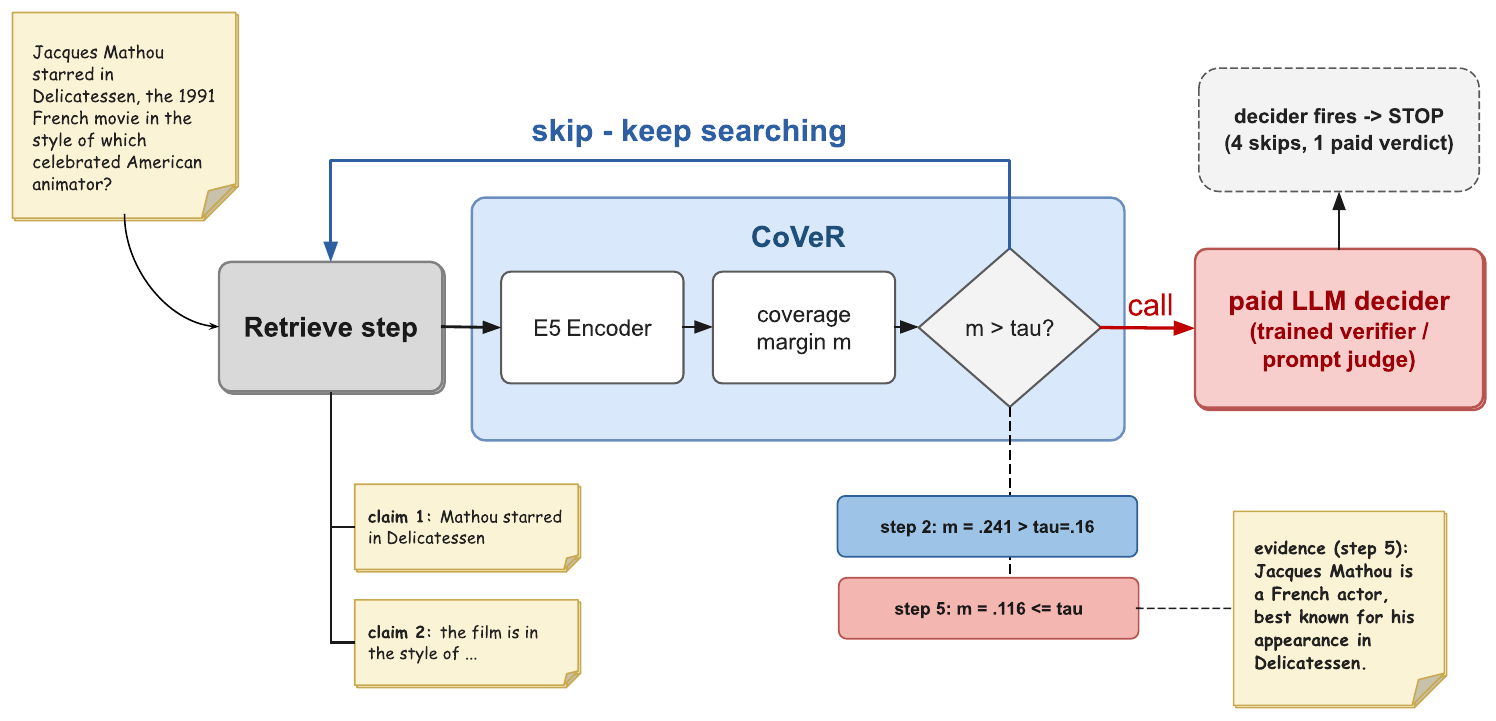}
\caption{\textbf{CoVeR routes the decider} (a routed HotpotQA trajectory).
Claims and evidence are embedded by a frozen encoder; one threshold on the
coverage margin skips the clearly-uncovered steps and pays for the one
ambiguous state, where the swappable decider rules. A skip can only defer
stopping --- every stop is decider-issued.
Source: lead-drawn figure (\texttt{figures/Figure.svg}, cropped).}
\label{fig:framework}
\end{figure}

\paragraph{The CoVeR gate.}
Figure~\ref{fig:framework} summarizes the gate and
Algorithm~\ref{alg:cover} gives the full loop: before each decider call,
evaluate the state margin:
\begin{align*}
  \cm(\ell)>\thigh&:\ \textbf{skip the call, continue};\\
  \cm(\ell)\le\thigh&:\ \textbf{call } D \text{ and follow its verdict.}
\end{align*}
We call this rule \textbf{CoVeR}. As in \S\ref{sec:intro}, a skip cannot end a
trajectory, so the rule is one-sided by construction.

One-sidedness is a design consequence, not a convenience. The margin is built
to detect states that are \emph{plainly incomplete}, not states that are
\emph{sufficient}: a large margin means no retrieved sentence comes close to
some expected claim, which is cheap evidence that the search is unfinished. The
converse does not follow. Establishing that evidence \emph{supports} a claim
requires reading how entities and relations compose, and cosine proximity does
not establish it. We therefore trust the margin only on the side where a
geometric check is reliable, and leave every stop to the decider.
Section~\ref{sec:results-replace} measures both sides of this asymmetry
directly. A second, lower early-exit threshold
(stop-without-a-call) would break the one-sided guarantee, so the primary
system uses the single threshold; we evaluate the two-threshold variant
only as an ablation (\S\ref{app:pooled}).

\begin{algorithm}[t]
\caption{The CoVeR-gated agent loop}
\label{alg:cover}
\begin{algorithmic}[1]
\REQUIRE question $q$; decider $D$; encoder $\phi$; threshold $\thigh$;
budget $L$
\STATE $\{e_j\}_{j=1}^{k}\leftarrow\phi(\mathrm{Claims}(q))$ or drafter$(q)$ \COMMENT{embedded LLM claim text, or predicted embeddings}
\STATE $S_0\leftarrow\emptyset$
\FOR{$\ell=1$ \TO $L$}
  \STATE $S_\ell\leftarrow S_{\ell-1}\cup\mathrm{retrieve}(q,\ell)$ \COMMENT{embed new sentences}
  \STATE $\cm(\ell)\leftarrow\max_j\bigl(1-\max_{s\in S_\ell}\cos(e_j,\phi(s))\bigr)$
  \IF{$\cm(\ell)>\thigh$}
    \STATE \textbf{continue} \COMMENT{skip: defer, never stop}
  \ENDIF
  \STATE $v\leftarrow D(S_\ell)$ \COMMENT{pay for a verdict}
  \IF{$v=$ all covered}
    \RETURN answer \COMMENT{stop is decider-issued}
  \ENDIF
\ENDFOR
\RETURN answer at budget exhaustion
\end{algorithmic}
\end{algorithm}

\paragraph{Threshold selection.}
Thresholds are selected on training splits only, leave-one-dataset-out
(LODO):
$\thigh$ for a target dataset is chosen on the \emph{other two} datasets'
train-$600$ and evaluated once on the target's held-out full split.
Selection maximizes verifier-call reduction subject to two training-side
constraints: \emph{stop-loop agreement} (the fraction of training questions
whose CoVeR-gated and always-verify runs stop at the same loop) must stay at
or above $0.90$, and \emph{coverage safety} (the fraction of the gate's
early, fired stops whose gold-labeled all-claims-covered loop has been
reached by that stop loop) must be at least the decider's own. This yields
$\thigh{=}0.16$ on all three folds.
Both bars are fixed across folds; only the threshold is searched, and only on
the other datasets' training splits. All three folds select the same value, so
the operating point does not depend on which dataset is held out.\ %% e2c

\paragraph{Extension: removing claim generation from the gate.}
The primary CoVeR system uses generated expected claims, costing one LLM
claim-generation pass per question. As a deployment extension we distill
those embeddings into a lightweight question-to-claim drafter. A sub-$1$M MLP maps the question embedding
to $k$ slot embeddings. It is trained with a cosine loss against the
generated-claim embeddings, while a small classification head predicts $k$;
there are \emph{no gold labels}. At inference the predicted embeddings replace
the generated claim text \emph{in the margin computation}, so the gate needs
only the frozen encoder and this MLP; a decider that consumes claim text
(the trained verifier) still requires claim generation when called --- a cost
always-verify pays too.

\paragraph{Extension: intra-call ordering.}
When $D$ is called, the all-covered check short-circuits at the first uncovered
claim. When scoring is sequential, $D$ may optionally evaluate claims in
\emph{descending}-margin order, which puts the most-likely-uncovered claim first, so a non-firing call reaches its first uncovered claim sooner.
Because the all-covered check is order-invariant, reordering preserves the
verdict exactly; the saving is realized only when claims are scored sequentially
(\S\ref{sec:results-eff}).

% ======================================================================
\section{Experimental Setup}
\label{sec:setup}

\paragraph{Datasets, agents, deciders.}
We evaluate on HotpotQA (distractor) \citep{yang2018hotpotqa}, 2WikiMultihopQA (2Wiki)
\citep{ho2020constructing}, and MuSiQue \citep{trivedi2022musique}. The host
agent is Self-Ask \citep{press2023selfask} with Qwen2.5-3B-Instruct
\citep{qwen2_5} (and Qwen2.5-7B for the scale test), maximum $6$ retrieval
loops (the simulator's budget parameter of $8$ was never binding), over a
BM25 sentence index \citep{robertson2009bm25} of gold$+$distractor paragraphs,
top-$K{=}3$ per query (a closed pool isolating stopping from retrieval failure).
We use \emph{two deciders}: a public \textbf{zero-shot prompt-only
sufficiency judge} (same Qwen2.5-3B, no training, stop iff $p_{\text{yes}}\ge
\tau_{\text{judge}}$), and a \textbf{trained-verifier instantiation} (a
Qwen2.5-3B$+$LoRA coverage verifier \citep{hu2022lora}; provenance below).
Reporting both decouples the method from any model with unreleased weights. The primary
statistics (\S\ref{sec:results-route}) are run on the trained verifier; the
prompt judge is an independent public replication under the same threshold
(\S\ref{sec:results-transfer}).

\paragraph{Systems.} We compare three systems built on identical trajectories:
the \textbf{full-budget} agent runs all $L$ retrieval loops; the
\textbf{always-verify} agent invokes the decider after every loop and stops when
it fires; the \textbf{CoVeR-gated} agent invokes the decider only when
$\cm(\ell)\le\thigh$. Full-budget is not assumed optimal; it is a conservative
no-early-stop reference that isolates whether gate-induced deferral sacrifices the
answer quality available under the retrieval budget.

\paragraph{Endpoints.}
EM is exact match, scored as Std-Ext EM: a frozen Qwen2.5-3B extractor reads cumulative chunks at
the stop loop, so all arms share the same extractor procedure. The endpoint
is $\Delta$EM of the CoVeR-gated agent against each reference: the
full-budget agent (a conservative reference, not an assumed optimum) and the
always-verify baseline (the stopping policy whose calls the gate
routes). \emph{Call cut}
is the fraction of the always-verify run's decider calls that the gate
skips on the same trajectories, and \emph{stop-loop agreement} is the fraction
of questions on which the gated and always-verify runs stop at the same loop.
Reported together, the two endpoints separate outcome preservation from policy
imitation: EM measures the answer the agent finally returns, while agreement
measures how closely the routed agent reproduces the decider's own stopping
behavior. The main result
evaluates $n{=}1000/1000/911$ unique questions on HotpotQA/2Wiki/MuSiQue
(populations differ by protocol; transfer arms use their own splits,
\S\ref{sec:results-transfer}), and $4$-bit re-extraction can flip
individual questions; full accounting in Appendix~\ref{app:denom}.

\paragraph{Confidence intervals.}
All CIs are
paired bootstrap ($10{,}000\times$ main, $2{,}000\times$ AUROC, seed $13$;
\citealp{efron1993bootstrap}); AUROC CIs cluster by question. With
$n{\le}1000$ at this EM level, effects below $\sim\!1$\,pp are within
noise on HotpotQA/2Wiki.\ %% stat02
A formal non-inferiority and equivalence analysis, with margins fixed before
the full run, supports the same conclusions and is reported in
Appendix~\ref{app:pooled}.\ %% stat02

\paragraph{Artifacts and protocol.}
The trained verifier is HALT \citep{roh2026halt}, our concurrent work; we use
its cached trajectories, verdicts, generated claims, and
extractions read-only, without modifying or retraining it --- the verifier
learns whether evidence supports expected claims; CoVeR neither changes that
training nor predicts its verdict, estimating only whether an invocation can
alter the current control action. Our code and archived result sets --- every
number in this paper traces to them --- will be released publicly upon
publication.
% Code release deliberately deferred (lead's decision 2026-08-04): the repo
% goes public after the review outcome. When it does, replace the sentence
% above with the public URL (https://github.com/Noverse0/CoVeR_code) and drop
% "will be released ... upon publication".
Every success bar was fixed before the corresponding full-scale run, and
freshly extracted arms are parity-checked against the cache ($\geq0.98$).
Complete provenance, parity checks, pilot analyses, and protocol history are
reported in Appendix~\ref{app:pooled}.

% ======================================================================
\section{Results}
\label{sec:results}

Stopping verification bundles two abilities: detecting that evidence is
still plainly incomplete, and judging that it has become sufficient --- and
our results test whether they separate, the first served by cheap geometry,
the second reserved for the decider. We report the routing
result first (\S\ref{sec:results-route}), then the asymmetry that licenses it
(\S\ref{sec:results-replace}), the comparison against cheaper routers
(\S\ref{sec:results-ctl}), transfer (\S\ref{sec:results-transfer}), and
workload (\S\ref{sec:results-eff}).

\subsection{Main result: answer quality preserved at a fraction of the verifier calls}
\label{sec:results-route}

Table~\ref{tab:main} reports the primary LODO result for CoVeR with the
trained-verifier decider, evaluated on up to $1000$ questions per dataset. The
CoVeR-gated agent matches both references within a fraction of an EM point
while cutting the majority of verifier calls.

\begin{table*}[t]
\centering\small
\resizebox{\textwidth}{!}{%
\begin{tabular}{lcccccrrrr}
\toprule
 & \multicolumn{3}{c}{EM} & \multicolumn{2}{c}{$\Delta$EM (pp) vs.} & Call cut & Loop $\Delta$ & Agree.\ \\
\cmidrule(lr){2-4}\cmidrule(lr){5-6}
Dataset & Full-budget & CoVeR & Always-verify & Full-budget & Always-verify & vs.\ Always & vs.\ Full & vs.\ Always \\
\midrule
HotpotQA & 0.3060 & 0.3040 & 0.3050 & $-0.20$ & $-0.10$ & $62.2\%$ & $-11.9\%$ & $.910$ \\ % stat02
2Wiki    & 0.1840 & 0.1870 & 0.1850 & $+0.30$ & $+0.20$ & $67.8\%$ & $-9.8\%$ & $.886$ \\ % stat02
MuSiQue$^\dagger$ & 0.0801 & 0.0812 & 0.0812 & $+0.11$ & $+0.00$ & $93.0\%$ & $-0.6\%$ & $.976$ \\ % stat02
\bottomrule
\end{tabular}}
\caption{\textbf{Routing at up to $n{=}1000$, LODO thresholds,
trained-verifier decider.} $\Delta$EM in pp (full $95\%$ CIs in
Table~\ref{tab:ci}; formal tests in Appendix~\ref{app:pooled}). Agree.\ = held-out stop-loop agreement ($\geq0.90$
applies at LODO selection). $^\dagger$Saturated (\S\ref{sec:results-route}).
Source: \texttt{results/stat02/}.}
\label{tab:main}
\end{table*}

Across the three datasets, EM differs from the full-budget and
always-verify references by at most $0.3$\,pp, while verifier calls fall by
$62.2\%$ on HotpotQA and $67.8\%$ on 2Wiki; every $95\%$ CI on these
differences lies within $\pm1.2$\,pp (the full intervals, a pooled
analysis, and the previously specified equivalence tests are in
Appendix~\ref{app:pooled}).\ %% stat02
A post-hoc token-F1 check shows the same pattern: Full/CoVeR/always-verify
$.468/.463/.462$, $.292/.294/.291$, $.164/.165/.166$ (secondary metric;
Appendix~\ref{app:pooled}).\ %% stat02
MuSiQue is a saturated regime and is interpreted separately below.\ %% stat02
The loop column reads in the gate's favor as well: the gated agent finishes
$9.8$--$11.9\%$ \emph{fewer} retrieval loops than the full-budget agent on
HotpotQA/2Wiki --- adjudicated stops end trajectories the budget reference
would run to exhaustion --- while deferring only mildly relative to
always-verify (\S\ref{sec:results-eff}).\ %% stat02

\paragraph{Three datasets, two regimes.}
The three rows are not three replications of one effect; they differ in how
much room the decider leaves to route. HotpotQA and 2Wiki are \emph{headroom}
regimes: the verifier fires on a substantial share of loops, so the calls it
receives include many whose verdict is genuinely in doubt, and skipping only
the plainly incomplete states removes about two thirds of the verifier calls
while preserving answer accuracy. MuSiQue is \emph{saturated}: the verifier
fires on $2.6\%$ of loops, most trajectories reach the loop budget under every
system, and the gate can skip almost every call precisely because almost none
of them would have changed the control action.\ %% stat02
The saturated cell therefore tests safety rather than savings --- near-total
skipping does not break a decider that was not going to stop anyway --- which
is why we report $62$--$68\%$ as the headline reduction: it is the range
measured where the stopping decision was actually live.

\subsection{Why the margin can route but cannot replace}
\label{sec:results-replace}

This subsection tests the design premise of \S\ref{sec:method}: identifying
plainly incomplete states is enough to route, but not enough to decide that a
state has become sufficient. Table~\ref{tab:oracle}
reports how well the margin separates covered from uncovered claims against
per-hop verifier labels, with clustered bootstrap CIs. With oracle targets the
margin reaches AUROC $.89/.90/.83$; the deployable
generated-claim margin reaches only $.69/.68/.76$, a gap whose CIs do not overlap.\ %% stat01
A label-free sub-$1$M predictor does not close this gap: it lands at or
below the generated-claim floor on every dataset, because its only available
training target (the rank-$0$ retrieved chunk) teaches the retrieval
distribution, not which evidence is \emph{needed}.\ %% e1
Turning the deployable margin into a stopping rule confirms the
consequence: tuned to match the verifier's coverage-safety, the threshold drives
loop savings to $\approx0$ and the rule never stops
on any dataset at a safety-matched operating point.\ %% e2
Deciding that evidence supports a claim requires reading how entities and
relations compose (a distinction NLI and attribution evaluation have long
enforced \citep{thorne2018fever,gao2023alce}), and cosine proximity does not
cross it. This is why CoVeR retains the decider for every stopping
decision.

\begin{table}[t]
\centering\small
\begin{tabular}{lcc}
\toprule
Dataset & Oracle AUROC (CI) & Generated-claim AUROC (CI) \\
\midrule
HotpotQA & $0.892\ (0.879,0.904)$ & $0.686\ (0.659,0.713)$ \\ % stat01
2Wiki    & $0.895\ (0.884,0.905)$ & $0.680\ (0.655,0.703)$ \\ % stat01
MuSiQue  & $0.831\ (0.812,0.849)$ & $0.760\ (0.723,0.792)$ \\ % stat01
\bottomrule
\end{tabular}
\caption{\textbf{Margin vs.\ per-hop coverage ($n{=}1000$; clustered bootstrap
CIs).} A high oracle ceiling with a much weaker deployable
floor (non-overlapping CIs): the geometry can express coverage, but matching a
generated claim string cannot read it. Oracle and generated-claim targets come
from different label
runs/positive rates (ceiling/floor contrast, not a controlled ablation).
Sources: \texttt{results/gate01,stat01/}.}
\label{tab:oracle}
\end{table}

\subsection{Comparison with simpler routing signals}
\label{sec:results-ctl}

Would a cheaper signal route as well? We compare the frozen-embedding margin
against three same-interface controls on a shared frontier: an
\emph{evidence-count} router (skip while few sentences are retrieved), a
\emph{lexical-overlap} router (skip while content-word recall of each claim's
tokens is low), and a \emph{BM25} router (a per-claim Okapi-BM25 coverage margin
over the question's own pool). For each router we characterize the eval-set frontier: the maximum call cut it
reaches while holding stop-loop agreement with the always-verify decider at or
above $0.90$ (Table~\ref{tab:ctl}; LODO-selected operating points are in
\S\ref{app:controls2}).\ %% ctl01, def01

\begin{table}[t]
\centering\small
\begin{tabular}{lrrrrr}
\toprule
Dataset & Embedding & Count & Lexical & BM25 & Learned \\
\midrule
HotpotQA & \textbf{0.622} & 0.000 & 0.512 & 0.202 & 0.435 \\ % ctl01, def01, ctl05
2Wiki    & \textbf{0.623} & 0.288 & 0.562 & 0.191 & 0.563 \\ % ctl01, def01, ctl05
MuSiQue  & \textbf{1.000} & 0.999 & 0.999 & 0.339 & \textbf{1.000} \\ % ctl01, def01, ctl05
\bottomrule
\end{tabular}
\caption{\textbf{Cheaper-router control: max call cut at stop-agreement
$\geq0.90$} (best in bold; MuSiQue saturated). BM25: per-question closed-pool,
pool-max normalized; Learned: LR on the cheap signals (\S\ref{app:controls2}).
Sources: \texttt{results/ctl01,def01,ctl05/}.}
\label{tab:ctl}
\end{table}

\begin{figure}[t]
\centering
\includegraphics[width=\linewidth]{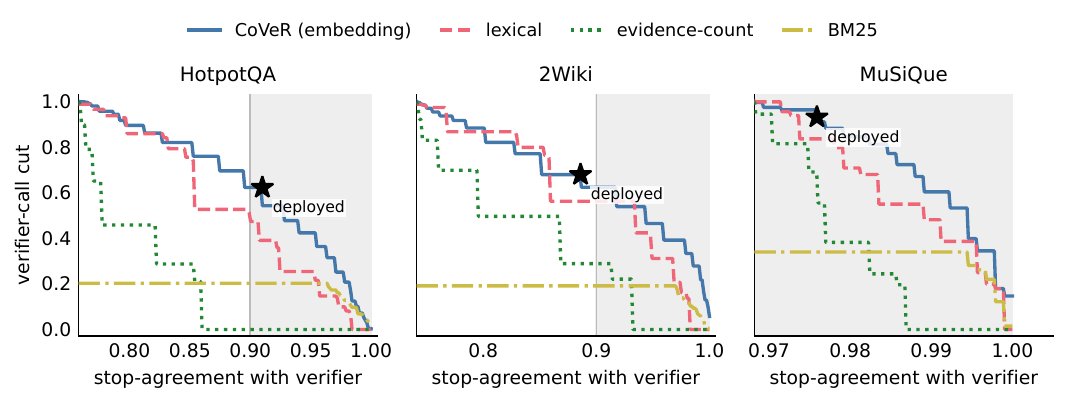}
\caption{\textbf{Router frontier (descriptive).} Achievable verifier-call cut
vs.\ stop-agreement with the always-verify decider, per dataset (envelope: max call cut
at agreement $\geq$ each $x$). Shaded region: agreement $\geq0.90$, the
pre-specified safety bar; the star marks the primary one-threshold operating
point (call cut $0.622$ at agreement $0.910$ on HotpotQA; STAT-02). Per-dataset readings are in the
text. Source: \texttt{results/ctl01/}.}
\label{fig:frontier}
\end{figure}

On this frontier the embedding provides the highest feasible cut at the
$0.90$ bar --- decisively on HotpotQA, modestly on 2Wiki, in a saturated tie
on MuSiQue --- and no cheaper proxy or learned combination of them exceeds it
(Table~\ref{tab:ctl}, Learned; \S\ref{app:controls2});
Table~\ref{tab:main}'s deployed point is the LODO-selected operating point,
not this frontier.
Matched-budget random skipping and verifier self-gating also fall short. A
random skipper calibrated to CoVeR's exact call cut loses $6.4$/$7.0$\,pp of
stop-agreement on HotpotQA/2Wiki (saturated MuSiQue ties; the pre-specified
$\geq5$\,pp bar is met on two of the three datasets), so the budget alone does not buy
the agreement --- the margin's choice of \emph{which} loops to skip
does.\ %% ctl03
And letting the verifier gate itself from its own unmatched-claim count reaches
an envelope below CoVeR's on every dataset at the $0.90$ agreement bar,
below CoVeR on all three datasets: self-gating pays a verifier call to obtain its
signal, so it cannot skip ahead of the state it has already paid to see
(Appendix~\ref{app:controls2}).\ %% ctl04
The evidence-count and BM25 routers are never competitive; against the
harder lexical router, the embedding's cut is $+10$--$17$\,pp larger on
HotpotQA at every agreement bar, matched on 2Wiki only at the $0.92$ bar,
and tied on saturated MuSiQue --- the advantage is largest where call-cut
headroom exists, and the mechanism behind it remains unresolved
(pre-specified tests; Appendix~\ref{app:mechanisms}). The single-threshold
frontier also reproduces the deployed operating point (call cut $0.622$ at
agreement $0.910$; Figure~\ref{fig:frontier}).\ %% stat02
The MuSiQue tie is mechanical: when the verifier rarely fires, most
trajectories end at the loop budget under always-verify as well, so skipping
calls cannot change their stop loop, and near-total cuts are open to any
signal --- though BM25's ordering still fails to protect the few
early-fired states (Table~\ref{tab:ctl}).\ %% ctl01
Read descriptively, the controls establish that the embedding is not
substitutable by a cheaper monotone coverage proxy at the operating point.
The advantage also survives an encoder audit: across five off-the-shelf
frozen encoders the deployed e5-base is strongest --- a twice-larger
e5-large only ties its frontier ($-3.1/{+}1.6$\,pp) while the bge/gte
retrieval families lose $12$--$25$\,pp of call cut --- and max aggregation
over evidence sentences beats mean and softmax variants
(Appendix~\ref{app:enc}).\ %% enc01

Routing is also not a repackaged loop-index rule: against skip-first-$k$ and
every-other-loop heuristics matched on either call cut or agreement, none
matches the gate on both (HotpotQA $0.63$ cut at $0.915$ agreement vs.\
skip-first-$4$'s $0.665$ at $0.762$).\ %% diag01
Two pre-specified lexical mechanisms --- a precision-side distractor-trap
statistic and a recall-side paraphrase-distance hypothesis --- both reproduce
the aggregate ordering the story predicts but fail per question
(Appendix~\ref{app:mechanisms}). CoVeR's benefit therefore depends on
identifying specific skippable states rather than on matching an overall skip
rate. The precise source of its advantage over lexical overlap remains
unresolved.\ %% ctl02, imp01

% In a two-column layout, promote to figure* (spans both columns).

\subsection{Transfer checks: decider, scale, corpus, and LLM-free routing}
\label{sec:results-transfer}

\paragraph{A public prompt judge as an independent decider.}
The trained verifier is one instantiation; the gate routes a decider that
is fully public. Replacing the in-band decider with a zero-shot
prompt-only sufficiency judge and keeping the \emph{same} threshold with no
re-tuning, all $6$ cells ($3$ datasets $\times$ $2$ judge thresholds) land within
$1$\,pp of the always-verify judge at $\geq40\%$ judge-call cut (cuts $56.7$--$95.2\%$).\ %% gen01
The judge's own absolute EM is lower than the trained verifier's, but it is a
fully public decider, and the gate preserves its accuracy while removing most
of its calls. These results indicate the routing behavior is not specific to
the trained verifier.
A fully public, off-the-shelf NLI cross-encoder (DeBERTa-v3 MNLI class; no
task-specific fine-tuning by us) over a declarative form of the same generated
claims is a third decider.
With the \emph{same} untuned threshold the gate removes $61$--$68\%$ of its calls
on HotpotQA/2Wiki and $94$--$95\%$ on MuSiQue, with $\Delta$EM within margin on
2Wiki/MuSiQue and strictly positive on HotpotQA ($+2.8$\,pp, CI above zero),
where the gate vetoes the decider's premature stops; a first, mechanically
invalid templated-hypothesis variant was degenerate and is reported alongside
(Table~\ref{tab:nli}).\ %% gen03, gen03b

\paragraph{A larger agent and a gate-side embedding drafter.}
Transferring the same threshold onto Self-Ask 7B trajectories (transfer arms
evaluate on the gold-order last-$400$ per dataset; MuSiQue $384$ unique), now
with the verifier on generated expected claims (the fully deployable
setting), again removes about two thirds of
verifier calls while EM stays within a fraction of a point of the always-verify baseline on
every dataset, with differences that are not statistically distinguishable
(Table~\ref{tab:transfer}a); the MuSiQue cell is a saturation case, as in
the 3B run. The behavior is not specific to the 3B agent, and it holds without gold
labels.\ %% def01
Replacing claim generation \emph{for the margin} with the sub-$1$M distilled
drafter (CoVeR-D) routes end-to-end and preserves the
generated-claim gate's EM at a comparable call cut
(Table~\ref{tab:transfer}b) --- the distilled gate preserves end-to-end
answer quality, not only per-claim discrimination.\ %% rev01
Pairing the drafter with the claim-free prompt judge removes claim
generation from the \emph{whole} pipeline: no claim text anywhere, and the
gate still routes the judge with EM within $1$\,pp of always-verify (CIs
containing $0$) at $56$--$92\%$ judge-call cut, on all three datasets and
at both judge thresholds (App.~\ref{app:claimsources}).\ %% rev02

\paragraph{Verifier firing rates drop under generated claims.}
Moving from the disclosed gold-claim decider to the deployable generated-claim
setting on the $7$B trajectories collapses the verifier's fire rate on every
dataset ($51.0\to19.2$ / $36.6\to14.0$ / $23.8\to2.6\%$)
(App.~\ref{app:claimsources} summarizes which claim source each arm uses).\ %% def01
Generated claims are harder to satisfy than gold supporting facts, so the
deployable verifier fires rarely on the larger agent; the gate's savings are
therefore largest precisely where verification rarely triggers: the gate acts
as a per-state detector of verifier usefulness, paying for verdicts only where
they can matter. EM still tracks the always-verify baseline (Table~\ref{tab:transfer}a), so this
is a lower-fire operating regime, not a degradation.

\paragraph{An initial open-corpus check.}
The main results use a closed distractor pool. Transferring the same untuned
threshold to an \emph{open-corpus} setting (HotpotQA Setting-B, Self-Ask 3B over
E5-dense Wikipedia-2018, $n{=}1000$; Table~\ref{tab:transfer}c) cuts $92.1\%$
of verifier calls at EM statistically indistinguishable from both Full and the always-verify baseline ($+0.30$\,pp each), stop-agreement
$0.97$.\ %% gen02
This is a saturation regime: the gold verifier fires on only $6.4\%$ of questions
(Full EM $0.109$), so in a low-coverage open corpus the
gate is a large call saver in this setting, with no detected loss in EM, rather than an accuracy lever, but the
routing mechanism, and the closed-pool threshold, transfer out of the distractor
pool without re-tuning. This arm uses the gold-claim verifier as decider (the
only one run against Setting-B, disclosed) and is a single open-corpus point.

\begin{table}[t]
\centering\small
\textbf{(a) $7$B-agent routing (deployable generated-claim decider)}\\[2pt]
\begin{tabular}{lrccrr}
\toprule
Dataset & $n$ & CoVeR & Always-verify & $\Delta$EM & Call cut \\
\midrule
HotpotQA & 959 & 0.2690 & 0.2669 & $+0.21$ & $64.7\%$ \\ % def01
2Wiki    & 996 & 0.1506 & 0.1506 & $+0.00$ & $69.6\%$ \\ % def01
MuSiQue$^\dagger$ & 871 & 0.0700 & 0.0700 & $+0.00$ & $94.4\%$ \\ % def01
\bottomrule
\end{tabular}\\[6pt]
\textbf{(b) Distilled sub-$1$M drafter (gate-side, LLM-free margin)}\\[2pt]
\begin{tabular}{lrccrr}
\toprule
Dataset & $n$ & Claim-text & Distilled & $\Delta$EM & Call cut \\
\midrule
HotpotQA & 400 & 0.2825 & 0.2925 & $+1.00$ & $61.2\%$ \\ % rev01
2Wiki    & 400 & 0.1625 & 0.1650 & $+0.25$ & $65.8\%$ \\ % rev01
MuSiQue  & 384 & 0.0625 & 0.0625 & $+0.00$ & $91.5\%$ \\ % rev01
\bottomrule
\end{tabular}\\[6pt]
\textbf{(c) Open-corpus (Setting-B) routing, 3B, gold-claim decider}\\[2pt]
\begin{tabular}{lrccrr}
\toprule
Dataset & $n$ & CoVeR & Always-verify & $\Delta$EM & Call cut \\
\midrule
HotpotQA & 1000 & 0.1120 & 0.1090 & $+0.30$ & $92.1\%$ \\ % gen02
\bottomrule
\end{tabular}
\caption{\textbf{Transfer.} (a) Same threshold on a $7$B agent, no re-tuning, with
the fully deployable \emph{generated-claim} in-band decider (DEF-01);
the always-verify baseline is the generated-claims always-verify arm; all three $\Delta$EM CIs
contain $0$ ($+0.21$ / $+0.00$ / $+0.00$
degenerate). \textbf{$^\dagger$MuSiQue is a saturation case} (generated-claim fire
$2.6\%$, zero-variance always-verify delta), a saturation case exactly
as the MuSiQue-3B main cell. (b) The $921$k-parameter distilled drafter replaces
the claim-generation LLM \emph{for the margin} at inference; the genclaims
decider still consumes claim text when called (REV-01). (c) The same untuned threshold in
an open-corpus (Wikipedia-2018 dense) setting, a $6.4\%$-fire saturation regime
with a gold-claim decider (GEN-02, disclosed). $\Delta$EM in pp vs.\ the panel's
reference. The earlier $7$B \emph{gold-claim} diagnostic (REV-01) read
$0.2690/0.1556/0.0802$ CoVeR at $64.4/69.5/94.0\%$ cut; it is retained as a
diagnostic only and superseded by panel (a). Sources:
\texttt{results/def01,rev01,gen02/}.}
\label{tab:transfer}
\end{table}

\paragraph{A public NLI cross-encoder as a third decider.}
The transfer arms above swap the agent and the claim source but keep the
HALT verifier as decider. A fully public, off-the-shelf NLI
cross-encoder (DeBERTa-v3 MNLI class, entailment probability) is a third decider
carrying no component with unreleased weights. A first attempt to feed it the verifier's
templated claim strings was degenerate---those strings are not natural
hypotheses, so entailment probability was $\approx0$ throughout; we
pre-specified a declarative reformulation of the same generated-claim fields as
a new input decision (premise $=$ evidence sentence; hypothesis $=$ a fixed
declarative template over the claim's title and target), recorded in the archived
result sets as GEN-03/GEN-03b. Table~\ref{tab:nli} reports the gated and pure
decider over this reformulation. With the same untuned threshold, the gate
removes $61$--$68\%$ of its calls on HotpotQA/2Wiki and $94$--$95\%$ on MuSiQue,
at $\Delta$EM within noise on 2Wiki/MuSiQue. On HotpotQA $\Delta$EM is strictly
positive ($+2.8$\,pp, CI above zero): the gate vetoes the decider's premature
stops---the same margin-override mechanism as for the prompt judge---so the
HotpotQA deviation is in the \emph{favorable}
direction rather than a regression. A symmetric fire-rate guard rejects as
trivial any cell whose decider fires on $<\!2\%$ or $>\!98\%$ of loops; every
cell here fires at $0.2$--$0.6$, so the routing comparison is informative. A
pre-declared keyphrase (SECONDARY) hypothesis reproduces the verdict, ruling out
a single-surface-form artifact.

\begin{table}[t]
\centering\footnotesize
\setlength{\tabcolsep}{5pt}
\textbf{(a) PRIMARY hypothesis (title\,$\leftrightarrow$\,target)}\\[2pt]
\begin{tabular}{lccccc}
\toprule
Dataset & $\theta$ & Gated/Pure & $\Delta$EM (CI) & Cut & Fire p/g \\
\midrule
HotpotQA & 0.5 & .260/.233 & $+2.8\,(1.0,4.8)$   & $63\%$ & .52/.28 \\ % gen03b
HotpotQA & 0.9 & .278/.258 & $+2.0\,(0.5,3.8)$   & $61\%$ & .43/.23 \\ % gen03b
2Wiki    & 0.5 & .152/.140 & $+1.3\,(0.0,2.5)$   & $68\%$ & .61/.24 \\ % gen03b
2Wiki    & 0.9 & .150/.142 & $+0.8\,(-0.3,2.0)$  & $67\%$ & .45/.18 \\ % gen03b
MuSiQue  & 0.5 & .068/.057 & $+1.0\,(0.0,2.3)$   & $95\%$ & .40/.03 \\ % gen03b
MuSiQue  & 0.9 & .068/.062 & $+0.5\,(-0.5,1.6)$  & $94\%$ & .23/.02 \\ % gen03b
\bottomrule
\end{tabular}\\[6pt]
\textbf{(b) SECONDARY (keyphrase) hypothesis}\\[2pt]
\begin{tabular}{lccccc}
\toprule
Dataset & $\theta$ & Gated/Pure & $\Delta$EM (CI) & Cut & Fire p/g \\
\midrule
HotpotQA & 0.5 & .280/.260 & $+2.0\,(0.8,3.5)$   & $62\%$ & .40/.22 \\ % gen03b
HotpotQA & 0.9 & .285/.273 & $+1.3\,(0.3,2.5)$   & $61\%$ & .28/.16 \\ % gen03b
2Wiki    & 0.5 & .155/.155 & $+0.0\,(-1.8,1.8)$  & $67\%$ & .47/.18 \\ % gen03b
2Wiki    & 0.9 & .160/.163 & $-0.3\,(-1.5,1.0)$  & $68\%$ & .32/.15 \\ % gen03b
MuSiQue  & 0.5 & .065/.060 & $+0.5\,(-0.5,1.6)$  & $94\%$ & .35/.02 \\ % gen03b
MuSiQue  & 0.9 & .065/.057 & $+0.8\,(0.0,1.8)$   & $95\%$ & .21/.02 \\ % gen03b
\bottomrule
\end{tabular}
\caption{\textbf{Public NLI cross-encoder as a third decider (GEN-03b).}
DeBERTa-v3 MNLI-class entailment over a declarative reformulation of the
generated claims, gated by the same untuned threshold ($\thigh{=}0.16$) on the
coverage margin. Gated/Pure $=$ EM with/without the gate; $\Delta$EM in pp
(gated$-$pure) with $95\%$ paired-bootstrap CI; Cut $=$ decider-call reduction;
Fire p/g $=$ decider fire rate pure/gated. Every cell fires at $0.2$--$0.6$
(non-degenerate). HotpotQA $\Delta$EM is positive with CI above zero (the gate
vetoes premature stops); 2Wiki/MuSiQue are within margin. Eval
last-$400$/dataset (MuSiQue unique-qid). Source: \texttt{results/gen03b/}.}
\label{tab:nli}
\end{table}

\subsection{Verification workload}
\label{sec:results-eff}

\paragraph{Two tiers, two regimes.}
We report verifier workload at three levels --- whole decider calls,
claim-level evaluations, and input tokens --- and account for the middle level
as claim-verdict \emph{pairs} --- one claim adjudicated once --- per question
(Figure~\ref{fig:computestack}; per-stage counts in Appendix~\ref{app:eff}).
Tier 1, call routing, removes whole decider calls and holds whether the decider
is batched or not, eliminating $62$--$68\%$ of pairs on HotpotQA and 2Wiki
and $93\%$ on saturated MuSiQue. Tier 2, intra-call ordering, removes claims
within a retained call; it forgoes intra-call batching, so its additional
saving is realized only when claims are scored sequentially. Stacked, the two
tiers remove $76$--$96\%$ of pairs.\ %% eff01

\begin{figure}[t]
\centering
\includegraphics[width=0.85\linewidth]{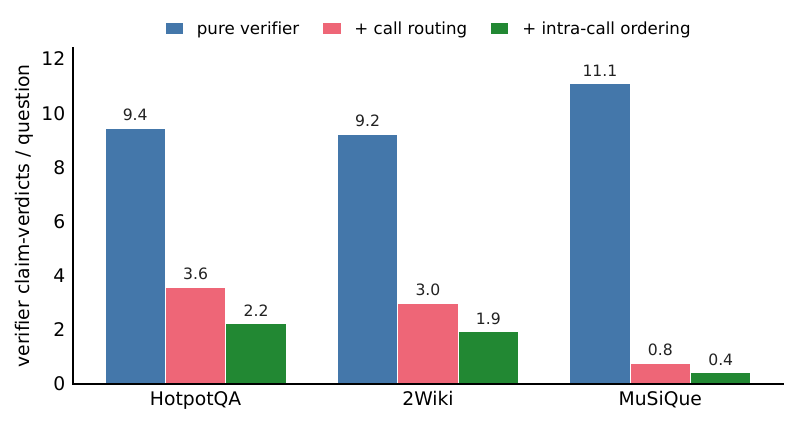}
\caption{\textbf{Verification compute stack.} Verifier claim-verdicts
per question at each routing tier: pure verifier ($9.4/9.2/11.1$), $+$ tier-1
call routing ($3.1/3.0/0.8$), $+$ tier-2 intra-call ordering ($2.0/1.9/0.4$). Tier 1
holds batched; tier 2 saves only unbatched. Source: \texttt{results/eff01/}.}
\label{fig:computestack}
\end{figure}

\paragraph{Workload and latency scope.}
The claim is a reduction in verifier \emph{workload}, in deployment-portable
units: $62$--$68\%$ fewer decider calls, $62$--$68\%$ fewer claim-verdict
pairs, and $72/73/97\%$ fewer verifier input tokens (a GPU replay confirms
the token and wall-clock reductions; Appendix~\ref{app:eff}). Gating also
defers stops, adding at most half a retrieval loop per question vs.\
always-verify; whether the saved verifier work outweighs the added loops
depends on the relative cost of a verifier call and a generation loop in a
given deployment --- we make no deployment-independent end-to-end latency claim, and give a
break-even rule plus one stack-specific timing study in
Appendix~\ref{app:eff}.
(\emph{Drafter cost.}) The generated-claim gate spends one claim-generation
LLM pass per question --- a pass always-verify also pays; the distilled
drafter replaces it for the margin computation (a sub-$1$M MLP forward).

The margin can also drive the symmetric decision of which native-stopped
trajectories to resume; that bounded positive result is reported in
Appendix~\ref{app:continue}.
In a new deployment the recipe is: embed the claims the agent already
generates, select the single threshold on a small training split
(\S\ref{sec:method}), and route; the same LODO-selected $\thigh{=}0.16$
transferred across deciders and agent scales, as well as to the open-corpus
setting.

% ======================================================================
\section{Analysis: When Can the Routing Signal Correct the Decider?}
\label{sec:analysis}

Routing and correction impose different requirements on the cheap signal:
routing needs only calibration (\S\ref{sec:related}), while correction
requires error decorrelation
\citep{madras2018predict,mozannar2020consistent}. Our two deciders
instantiate the two sides of that condition, and the contrast explains the
asymmetry.

\subsection{Correlated errors preclude correction: the trained verifier}

We partition questions into three paths: those that stop after a retained decider
call, those that reach the retrieval budget, and those whose trajectory diverges
after a skipped call.
Decomposing the gate's $\Delta$EM-vs-Full by path (primary arm, full $n$),
skip-diverged trajectories differ by at most one question and the budget
path equals Full up to re-extraction flips; the substantive residual is concentrated in the
\emph{decider-stop} path
(HotpotQA: $155$ questions, $-1.94$\,pp on that subset), and it is inherited
from always-verify \emph{exactly} ($\Delta$EM-vs-always $=0$ on that path):
early wrong stops fire near loop $1.3$ and match a wrong-entity page at rate
$0.68$ vs.\ $0.52$ for correct stops (on 2Wiki the same path inherits a small
gain, $+2.4$\,pp).\ %% diag01b
So the correction question is whether the margin can veto \emph{those} verifier
mistakes.

A pre-specified veto gate (evaluated on the $n{=}400$ pilot arm) asks whether
the margin-at-fire, or the verifier's own match-margin, ranks error fires
above correct fires at AUROC $\geq0.65$. It fails. The margin-at-fire AUROC is $0.39$ on the decider-stop
path and $0.36$ over all verifier fires; both confidence intervals include
chance-level AUROC of $0.5$,
so the margin gives \emph{no recoverable veto signal} (primary-arm error
counts stay small, $14/6/1$ --- the direction is not resolvable, only the
absence of a usable signal is).\ %% stat01, diag02, diag01b
Only a trivial fire-loop control separates ($0.76$), which largely reduces to a
position-based skip-first-$k$ heuristic. The reason is mechanistic: margin and
verifier are \emph{both} driven by claim--evidence similarity, so evidence
deceptive enough to fool the verifier also looks covered to the margin ---
their errors correlate, exactly the non-complementary regime where deferral
theory predicts no gain. The inherited decider-stop residual
remains a limit of the method.

\subsection{Decorrelated errors may permit it: the volatile judge (exploratory)}

The prompt judge shows a different error pattern. Its per-loop $p_{\text{yes}}$ is volatile (mean
absolute loop-to-loop change $0.09$--$0.16$), so its early ``sufficient''
calls are unreliable \emph{and} margin-separable; when it stops early at a
high-margin loop the gate vetoes that stop. Across all such overrides the effect is positive in direction but underpowered:
pooled $+1.77$\,pp at $\tau_{\text{judge}}{=}0.5$, with a CI spanning zero,
and similar at $0.95$.\ %% stat01
That instability leaves the margin room to veto an unstable early ``sufficient''
at a high-margin loop, unlike the stable trained verifier, whose shared failure
mode leaves no such room.
The NLI decider shows the same override pattern with a CI above zero
(\S\ref{sec:results-transfer}). We therefore treat the judge results as evidence consistent with, rather than
establishing, the decorrelation hypothesis. What the data does separate cleanly
is the two deciders: the judge's errors are visible to the margin, the trained
verifier's are not: the learning-to-defer complementarity condition,
instantiated for RAG coverage verification.

\section{Limitations}
\label{sec:limitations}

\paragraph{Scope.}
Our evidence is one agent scaffold (Self-Ask) and three multi-hop QA datasets;
the closed-pool limitation is only partially addressed by one open-corpus point
(HotpotQA Setting-B, \S\ref{sec:results-transfer}), a single saturated-regime
dataset with a gold-claim decider, and broader open-corpus routing and non-QA
tasks remain untested. Pilot and full-run populations overlap; the
deployable pipeline inherits generated-claim quality; the trained verifier's
weights belong to HALT (we disclose its verdicts and
spec; the weights are not ours to release). Three further threats: the frozen E5 encoder was
pretrained on Wikipedia-heavy corpora, so the margin may be weaker off-Wikipedia;
every trajectory is a single seed-13 run, so the bootstrap captures question-level
but not generation randomness; and claim decomposition itself can fail on
comparison or aggregation questions, a failure mode we do not quantify.

\paragraph{Generated-claim operation at 7B.}
The $7$B arm runs the verifier in the generated-claim setting (fully deployable)
(Table~\ref{tab:transfer}a), which resolves the gold-claim caveat at
scale, but the deployable always-verify baseline fires rarely on the stronger agent
($19.2/14.0/2.6\%$), so its MuSiQue cell is again saturated and
uninformative. Setting-B still uses a gold-claim decider
(disclosed).

\paragraph{Statistical resolution.}
Absolute EM is low ($0.08$--$0.31$), so sub-$1$pp effects are unresolvable 
on
HotpotQA/2Wiki and the HotpotQA CI does not exclude a $1$\,pp loss. The gate inherits the decider's early-stop errors unchanged
(decider-stop-path $\Delta$EM-vs-always $=0$), and \S\ref{sec:analysis} shows
the margin cannot veto them on single-digit error counts; MuSiQue's $93\%$ cut is a saturated regime, not a
transferable multiplier.

\paragraph{Cost model.}
The efficiency result (\S\ref{sec:results-eff}) is a verifier-workload
reduction in calls, pairs, and tokens; gate-side compute
(encoder$+$drafter) is not counted, and end-to-end consequences depend on
the serving stack (break-even rule and a measured instantiation:
Appendix~\ref{app:eff}). CoVeR-D removes claim generation only for the
margin when the decider consumes claim text.

% ======================================================================
\section{Conclusion}
\label{sec:conclusion}

CoVeR routes verifier calls using a single threshold on a frozen coverage
margin, while every stopping decision is still issued by the decider. Across
three multi-hop QA benchmarks it preserves answer accuracy relative to both
references while reducing verifier calls by $62$--$68\%$ in the representative
regimes. The same threshold, never tuned on the dataset it is evaluated on,
transfers across three deciders (a trained verifier, a public prompt judge,
and a public NLI cross-encoder) and across agent scales, a sub-$1$M drafter
removes LLM computation from the routing path, and an initial HotpotQA
experiment suggests the setting extends to an open corpus. The margin cannot
replace verification, nor correct failures it shares with the decider, which
indicates that correction would require a signal whose errors decorrelate from
the decider's.

\bibliographystyle{plainnat}
\bibliography{references}

\begin{thebibliography}{38}
\providecommand{\natexlab}[1]{#1}
\providecommand{\url}[1]{\texttt{#1}}
\expandafter\ifx\csname urlstyle\endcsname\relax
  \providecommand{\doi}[1]{doi: #1}\else
  \providecommand{\doi}{doi: \begingroup \urlstyle{rm}\Url}\fi

\bibitem[Asai et~al.(2024)Asai, Wu, Wang, Sil, and Hajishirzi]{asai2024selfrag}
Akari Asai, Zeqiu Wu, Yizhong Wang, Avirup Sil, and Hannaneh Hajishirzi.
\newblock {Self-RAG}: Learning to retrieve, generate, and critique through self-reflection.
\newblock In \emph{International Conference on Learning Representations}, 2024.

\bibitem[Bohnet et~al.(2022)Bohnet, Tran, Verga, Aharoni, Andor, Soares, Ciaramita, Eisenstein, Ganchev, Herzig, Hui, Kwiatkowski, Ma, Ni, Saralegui, Schuster, Cohen, Collins, Das, Metzler, Petrov, and Webster]{bohnet2022autoais}
Bernd Bohnet, Vinh~Q. Tran, Pat Verga, Roee Aharoni, Daniel Andor, Livio~Baldini Soares, Massimiliano Ciaramita, Jacob Eisenstein, Kuzman Ganchev, Jonathan Herzig, Kai Hui, Tom Kwiatkowski, Ji~Ma, Jianmo Ni, Lierni~Sestorain Saralegui, Tal Schuster, William~W. Cohen, Michael Collins, Dipanjan Das, Donald Metzler, Slav Petrov, and Kellie Webster.
\newblock Attributed question answering: Evaluation and modeling for attributed large language models.
\newblock \emph{arXiv preprint arXiv:2212.08037}, 2022.

\bibitem[Chen et~al.(2023)Chen, Borgeaud, Irving, Lespiau, Sifre, and Jumper]{chen2023speculative}
Charlie Chen, Sebastian Borgeaud, Geoffrey Irving, Jean-Baptiste Lespiau, Laurent Sifre, and John Jumper.
\newblock Accelerating large language model decoding with speculative sampling.
\newblock \emph{arXiv preprint arXiv:2302.01318}, 2023.

\bibitem[Chen et~al.(2024)Chen, Zaharia, and Zou]{chen2023frugalgpt}
Lingjiao Chen, Matei Zaharia, and James Zou.
\newblock {FrugalGPT}: How to use large language models while reducing cost and improving performance.
\newblock \emph{Transactions on Machine Learning Research (TMLR)}, 2024.

\bibitem[Chen et~al.(2025)Chen, Sun, Li, Sun, Zhou, Zhu, Wang, Pan, Zhang, Chen, Yang, Zhou, and Chen]{chen2025research}
Mingyang Chen, Linzhuang Sun, Tianpeng Li, Haoze Sun, Yijie Zhou, Chenzheng Zhu, Haofen Wang, Jeff~Z. Pan, Wen Zhang, Huajun Chen, Fan Yang, Zenan Zhou, and Weipeng Chen.
\newblock Research: Learning to reason with search for llms via reinforcement learning.
\newblock In \emph{Advances in Neural Information Processing Systems (NeurIPS)}, 2025.

\bibitem[Efron and Tibshirani(1993)]{efron1993bootstrap}
Bradley Efron and Robert~J. Tibshirani.
\newblock \emph{An Introduction to the Bootstrap}.
\newblock Chapman \& Hall/CRC, 1993.

\bibitem[Gao et~al.(2023)Gao, Yen, Yu, and Chen]{gao2023alce}
Tianyu Gao, Howard Yen, Jiatong Yu, and Danqi Chen.
\newblock Enabling large language models to generate text with citations.
\newblock In \emph{Proceedings of the 2023 Conference on Empirical Methods in Natural Language Processing (EMNLP)}, pages 6465--6488, 2023.

\bibitem[Ho et~al.(2020)Ho, Nguyen, Sugawara, and Aizawa]{ho2020constructing}
Xanh Ho, Anh-Khoa~Duong Nguyen, Saku Sugawara, and Akiko Aizawa.
\newblock Constructing a multi-hop {QA} dataset for comprehensive evaluation of reasoning steps.
\newblock In \emph{Proceedings of the 28th International Conference on Computational Linguistics}, pages 6609--6625, 2020.

\bibitem[Holm(1979)]{holm1979simple}
Sture Holm.
\newblock A simple sequentially rejective multiple test procedure.
\newblock \emph{Scandinavian Journal of Statistics}, 6\penalty0 (2):\penalty0 65--70, 1979.

\bibitem[Hu et~al.(2022)Hu, Shen, Wallis, Allen-Zhu, Li, Wang, Wang, and Chen]{hu2022lora}
Edward~J. Hu, Yelong Shen, Phillip Wallis, Zeyuan Allen-Zhu, Yuanzhi Li, Shean Wang, Lu~Wang, and Weizhu Chen.
\newblock {LoRA}: Low-rank adaptation of large language models.
\newblock In \emph{International Conference on Learning Representations}, 2022.

\bibitem[Java et~al.(2026)Java, Koundinyan, Natarajan, and Sharma]{java2025frugalrag}
Abhinav Java, Srivathsan Koundinyan, Nagarajan Natarajan, and Amit Sharma.
\newblock {FrugalRAG}: Less is more in {RL} finetuning for multi-hop question answering.
\newblock In \emph{International Conference on Learning Representations (ICLR)}, 2026.

\bibitem[Jeong et~al.(2024)Jeong, Baek, Cho, Hwang, and Park]{jeong2024adaptiverag}
Soyeong Jeong, Jinheon Baek, Sukmin Cho, Sung~Ju Hwang, and Jong~C. Park.
\newblock Adaptive-{RAG}: Learning to adapt retrieval-augmented large language models through question complexity.
\newblock In \emph{Proceedings of the 2024 Conference of the North American Chapter of the Association for Computational Linguistics: Human Language Technologies}, pages 7036--7050, 2024.

\bibitem[Jiang et~al.(2023)Jiang, Xu, Gao, Sun, Liu, Dwivedi-Yu, Yang, Callan, and Neubig]{jiang2023active}
Zhengbao Jiang, Frank~F Xu, Luyu Gao, Zhiqing Sun, Qian Liu, Jane Dwivedi-Yu, Yiming Yang, Jamie Callan, and Graham Neubig.
\newblock Active retrieval augmented generation.
\newblock In \emph{Proceedings of the 2023 conference on empirical methods in natural language processing}, pages 7969--7992, 2023.

\bibitem[Jin et~al.(2025)Jin, Zeng, Yue, Yoon, Arik, Wang, Zamani, and Han]{jin2025searchr1}
Bowen Jin, Hansi Zeng, Zhenrui Yue, Jinsung Yoon, Sercan Arik, Dong Wang, Hamed Zamani, and Jiawei Han.
\newblock {Search-R1}: Training {LLM}s to reason and leverage search engines with reinforcement learning.
\newblock In \emph{Conference on Language Modeling (COLM)}, 2025.

\bibitem[Joren et~al.(2025)Joren, Zhang, Ferng, Juan, Taly, and Rashtchian]{joren2025sufficient}
Hailey Joren, Jianyi Zhang, Chun-Sung Ferng, Da-Cheng Juan, Ankur Taly, and Cyrus Rashtchian.
\newblock Sufficient context: A new lens on retrieval augmented generation systems.
\newblock In \emph{International Conference on Learning Representations (ICLR)}, 2025.

\bibitem[Leviathan et~al.(2023)Leviathan, Kalman, and Matias]{leviathan2023speculative}
Yaniv Leviathan, Matan Kalman, and Yossi Matias.
\newblock Fast inference from transformers via speculative decoding.
\newblock In \emph{Proceedings of the 40th International Conference on Machine Learning (ICML)}, volume 202 of \emph{PMLR}, pages 19274--19286, 2023.

\bibitem[Lewis et~al.(2020)Lewis, Perez, Piktus, Petroni, Karpukhin, Goyal, K{\"u}ttler, Lewis, Yih, Rockt{\"a}schel, Riedel, and Kiela]{lewis2020rag}
Patrick Lewis, Ethan Perez, Aleksandra Piktus, Fabio Petroni, Vladimir Karpukhin, Naman Goyal, Heinrich K{\"u}ttler, Mike Lewis, Wen-tau Yih, Tim Rockt{\"a}schel, Sebastian Riedel, and Douwe Kiela.
\newblock Retrieval-augmented generation for knowledge-intensive {NLP} tasks.
\newblock In \emph{Advances in Neural Information Processing Systems}, volume~33, pages 9459--9474, 2020.

\bibitem[Li et~al.(2026)Li, Yan, and K{\"a}fer]{li2026raser}
Yuyang Li, Zihe Yan, and Tobias K{\"a}fer.
\newblock {RASER}: Recoverability-aware selective escalation router for multi-hop question answering.
\newblock \emph{arXiv preprint arXiv:2606.02488}, 2026.

\bibitem[Li et~al.(2023)Li, Zhang, Zhang, Long, Xie, and Zhang]{li2023gte}
Zehan Li, Xin Zhang, Yanzhao Zhang, Dingkun Long, Pengjun Xie, and Meishan Zhang.
\newblock Towards general text embeddings with multi-stage contrastive learning, 2023.

\bibitem[Madras et~al.(2018)Madras, Pitassi, and Zemel]{madras2018predict}
David Madras, Toniann Pitassi, and Richard Zemel.
\newblock Predict responsibly: Improving fairness and accuracy by learning to defer.
\newblock In \emph{Advances in Neural Information Processing Systems (NeurIPS)}, 2018.

\bibitem[Mozannar and Sontag(2020)]{mozannar2020consistent}
Hussein Mozannar and David Sontag.
\newblock Consistent estimators for learning to defer to an expert.
\newblock In \emph{Proceedings of the 37th International Conference on Machine Learning (ICML)}, volume 119 of \emph{PMLR}, 2020.

\bibitem[Park et~al.(2025)Park, Cho, and Lee]{park2025stop}
Jaewan Park, Solbee Cho, and Jay-Yoon Lee.
\newblock Stop-rag: Value-based retrieval control for iterative rag.
\newblock In \emph{NeurIPS 2025 Workshop on Multi-Turn Interactions in LLMs (MTI-LLM)}, 2025.

\bibitem[Press et~al.(2023)Press, Zhang, Min, Schmidt, Smith, and Lewis]{press2023selfask}
Ofir Press, Muru Zhang, Sewon Min, Ludwig Schmidt, Noah~A. Smith, and Mike Lewis.
\newblock Measuring and narrowing the compositionality gap in language models.
\newblock In \emph{Findings of the Association for Computational Linguistics: EMNLP 2023}, pages 5687--5711, 2023.

\bibitem[Qu(2026)]{qu2026adaptive}
Shuhui Qu.
\newblock Adaptive test-time compute allocation via learned heuristics over categorical structure.
\newblock \emph{arXiv preprint arXiv:2602.03975}, 2026.

\bibitem[Robertson and Zaragoza(2009)]{robertson2009bm25}
Stephen Robertson and Hugo Zaragoza.
\newblock The probabilistic relevance framework: {BM25} and beyond.
\newblock \emph{Foundations and Trends in Information Retrieval}, 3\penalty0 (4):\penalty0 333--389, 2009.

\bibitem[Roh and Han(2026)]{roh2026halt}
Daeyoung Roh and Donghee Han.
\newblock {HALT}: Verification-aware stopping for retrieval-augmented search agents.
\newblock \emph{arXiv preprint arXiv:2608.02009}, 2026.

\bibitem[Schuirmann(1987)]{schuirmann1987comparison}
Donald~J. Schuirmann.
\newblock A comparison of the two one-sided tests procedure and the power approach for assessing the equivalence of average bioavailability.
\newblock \emph{Journal of Pharmacokinetics and Biopharmaceutics}, 15\penalty0 (6):\penalty0 657--680, 1987.

\bibitem[Thorne et~al.(2018)Thorne, Vlachos, Christodoulopoulos, and Mittal]{thorne2018fever}
James Thorne, Andreas Vlachos, Christos Christodoulopoulos, and Arpit Mittal.
\newblock {FEVER}: a large-scale dataset for fact extraction and {VER}ification.
\newblock In \emph{Proceedings of the 2018 Conference of the North American Chapter of the Association for Computational Linguistics: Human Language Technologies (NAACL-HLT)}, pages 809--819, 2018.

\bibitem[Trivedi et~al.(2022)Trivedi, Balasubramanian, Khot, and Sabharwal]{trivedi2022musique}
Harsh Trivedi, Niranjan Balasubramanian, Tushar Khot, and Ashish Sabharwal.
\newblock {MuSiQue}: Multihop questions via single-hop question composition.
\newblock \emph{Transactions of the Association for Computational Linguistics}, 10:\penalty0 539--554, 2022.

\bibitem[Trivedi et~al.(2023)Trivedi, Balasubramanian, Khot, and Sabharwal]{trivedi2023interleaving}
Harsh Trivedi, Niranjan Balasubramanian, Tushar Khot, and Ashish Sabharwal.
\newblock Interleaving retrieval with chain-of-thought reasoning for knowledge-intensive multi-step questions.
\newblock In \emph{Proceedings of the 61st annual meeting of the association for computational linguistics (volume 1: long papers)}, pages 10014--10037, 2023.

\bibitem[Wang et~al.(2022)Wang, Yang, Huang, Jiao, Yang, Jiang, Majumder, and Wei]{wang2022e5}
Liang Wang, Nan Yang, Xiaolong Huang, Binxing Jiao, Linjun Yang, Daxin Jiang, Rangan Majumder, and Furu Wei.
\newblock Text embeddings by weakly-supervised contrastive pre-training, 2022.

\bibitem[Wang et~al.(2026)Wang, Brahma, and Henao]{wang2026sage}
Sijia Wang, Dhanajit Brahma, and Ricardo Henao.
\newblock {SAGE}: A novelty gate for efficient memory evolution in agentic {LLM}s.
\newblock \emph{arXiv preprint arXiv:2605.30711}, 2026.

\bibitem[Wang et~al.(2025)Wang, Wang, Le, Zheng, Mishra, Perot, Zhang, Mattapalli, Taly, Shang, Lee, and Pfister]{wang2024speculativerag}
Zilong Wang, Zifeng Wang, Long Le, Huaixiu~Steven Zheng, Swaroop Mishra, Vincent Perot, Yuwei Zhang, Anush Mattapalli, Ankur Taly, Jingbo Shang, Chen-Yu Lee, and Tomas Pfister.
\newblock Speculative {RAG}: Enhancing retrieval augmented generation through drafting.
\newblock In \emph{International Conference on Learning Representations (ICLR)}, 2025.

\bibitem[Xiao et~al.(2024)Xiao, Liu, Zhang, Muennighoff, Lian, and Nie]{xiao2024cpack}
Shitao Xiao, Zheng Liu, Peitian Zhang, Niklas Muennighoff, Defu Lian, and Jian-Yun Nie.
\newblock {C-Pack}: Packed resources for general {Chinese} embeddings.
\newblock In \emph{Proceedings of the 47th International ACM SIGIR Conference on Research and Development in Information Retrieval}, 2024.

\bibitem[Yang et~al.(2024)Yang, Yang, Zhang, Hui, Zheng, Yu, Li, Liu, Huang, Wei, et~al.]{qwen2_5}
An~Yang, Baosong Yang, Beichen Zhang, Binyuan Hui, Bo~Zheng, Bowen Yu, Chengyuan Li, Dayiheng Liu, Fei Huang, Haoran Wei, et~al.
\newblock {Qwen2.5} technical report, 2024.

\bibitem[Yang et~al.(2025)Yang, Zeng, Rao, and Zhang]{yang2025simrag}
Diji Yang, Linda Zeng, Jinmeng Rao, and Yi~Zhang.
\newblock Knowing you don't know: Learning when to continue search in multi-round {RAG} through self-practicing.
\newblock In \emph{Proceedings of the 48th International ACM SIGIR Conference on Research and Development in Information Retrieval}, 2025.

\bibitem[Yang et~al.(2018)Yang, Qi, Zhang, Bengio, Cohen, Salakhutdinov, and Manning]{yang2018hotpotqa}
Zhilin Yang, Peng Qi, Saizheng Zhang, Yoshua Bengio, William Cohen, Ruslan Salakhutdinov, and Christopher~D. Manning.
\newblock {HotpotQA}: A dataset for diverse, explainable multi-hop question answering.
\newblock In \emph{Proceedings of the 2018 Conference on Empirical Methods in Natural Language Processing}, pages 2369--2380, 2018.

\bibitem[Yao et~al.(2023)Yao, Zhao, Yu, Du, Shafran, Narasimhan, and Cao]{yao2023react}
Shunyu Yao, Jeffrey Zhao, Dian Yu, Nan Du, Izhak Shafran, Karthik Narasimhan, and Yuan Cao.
\newblock {ReAct}: Synergizing reasoning and acting in language models.
\newblock In \emph{International Conference on Learning Representations (ICLR)}, 2023.

\end{thebibliography}

\appendix

% ======================================================================
\section{Case Study: One Routed Trajectory}
\label{app:case}

Table~\ref{tab:case} walks through the full trajectory behind the
introduction's example (2WikiMultihopQA qid \texttt{dev\_624}; Figure~1 of
the main paper). The question needs each film's director and death date:
\emph{Remember the Day} $\to$ Henry King (d.~1982) and \emph{Cast Up by the
Sea} $\to$ John Gavin (d.~1938). The per-claim margins are recomputed from
the archived caches and reproduce the cached state margins exactly.

\begin{table*}[t]
\centering\footnotesize
\setlength{\tabcolsep}{3pt}
\begin{tabular}{cp{2.0cm}p{4.0cm}ccl}
\toprule
Step & Follow-up query & Key retrieved evidence (new that step) & $\cm_{\text{RtD}}$/$\cm_{\text{CUbtS}}$ & $\cm(\ell)$ & Gate ($\thigh{=}0.16$) \\
\midrule
1 & director of \emph{Remember the Day} & ``Remember the Day is a 1941 film released by 20th Century Fox, directed by Henry King\ldots'' (also retrieves John Gavin's bio, 1875--1938, \emph{without} any link to the film) & .174/.178 & .178 & skip \\
2 & when did Henry King die & ``Henry King (January 24, 1886 -- June 29, 1982) was an American actor and film director.'' & .174/.178 & .178 & skip \\
3 & director of \emph{Cast Up by the Sea} & ``\textbf{Cast Up by the Sea is a 1916 Australian film directed by John Gavin.}'' & .134/.131 & .134 & \textbf{call} $\to$ decider fires \\
\bottomrule
\end{tabular}
\caption{\textbf{A routed trajectory (2WikiMultihopQA).} The
\emph{Remember the Day} side is covered by step 2, but nothing links
\emph{Cast Up by the Sea} to a director, so the state margin (the
least-covered claim) holds above the threshold and the gate skips two
verifier calls whose verdicts could only have been ``keep searching.'' When
the missing film--director link arrives at step 3, both per-claim margins
collapse, the gate pays for one verdict, and the decider fires. Full-budget
runs three further steps to the loop-6 cap; full-budget, always-verify, and
CoVeR all return the correct answer (\emph{Cast Up by the Sea}, EM $1$).
Per-claim margins recomputed from our archived caches; state margins match
the cached values exactly.}
\label{tab:case}
\end{table*}

Two things are visible here that the aggregate tables cannot show. First,
the routed loops are not borderline: the pool contains lexically tempting
distractors --- a progressive-metal album also titled \emph{Remember the
Day}, and the red-herring sentence ``The music director died before the
release of the film'' --- yet none of them moves the margin, and John
Gavin's own biography (retrieved at step 1) does not cover the claim until
a sentence actually links him to the film. Second, the division of labor is
exactly as designed: the margin never decides that the state is covered ---
when the linking evidence arrives, that judgment is paid for and made by
the decider.

% ======================================================================
\section{Pooled Power Analysis: Populations and Endpoints}
\label{app:pooled}

\paragraph{Previously specified non-inferiority and equivalence tests.}
The main text reports paired-bootstrap $95\%$ CIs; this appendix retains the
confirmatory analysis specified before the full-scale run. Non-inferiority
(NI) vs.\ the full-budget agent is one-sided: the lower $95\%$ bound of
$\Delta$EM-vs-Full must exceed $-\varepsilon$ ($\alpha{=}0.025$).
Equivalence vs.\ the always-verify baseline uses TOST
\citep{schuirmann1987comparison}: the $90\%$ CI of $\Delta$EM must lie
within $\pm\varepsilon$. The margin $\varepsilon{=}2$\,pp was fixed
before the full run (a design margin, near the per-cell minimum detectable
effect, MDE $1.42/1.23/0.54$\,pp), with sensitivity reported at
$\{1,1.5,2\}$\,pp; the six-test family ($3$ NI $+$ $3$ TOST; degenerate
cells dropped) is Holm-corrected at $0.05$ \citep{holm1979simple}. All five
non-degenerate tests reject (max adjusted $p{=}0.0002$);
Table~\ref{tab:stats} reports the cells; Table~\ref{tab:ci} gives the full
paired-bootstrap intervals behind the $\pm1.2$\,pp bound quoted above. HotpotQA's $1$\,pp NI margin is
the one open cell --- it sits below the per-cell MDE, a power statement
rather than an observed loss.

\begin{table}[t]
\centering\small
\begin{tabular}{lrccccc}
\toprule
 & $\Delta$EM & \multicolumn{3}{c}{NI vs.\ Full ($\varepsilon$)} & TOST & MDE \\
\cmidrule(lr){3-5}
Dataset & vs.\ Full & $1$ & $1.5$ & $2$ & vs.\ alw-verify & (pp) \\
\midrule
HotpotQA & $-0.20$ & $\times$ & \checkmark & \checkmark & $\pm1,1.5,2$ & $1.42$ \\ % stat02
2Wiki    & $+0.30$ & \checkmark & \checkmark & \checkmark & $\pm1,1.5,2$ & $1.23$ \\ % stat02
MuSiQue  & $+0.11$ & \checkmark & \checkmark & \checkmark & degen.\ & $0.54$ \\ % stat02
\bottomrule
\end{tabular}
\caption{\textbf{Confirmatory statistics.} NI is one-sided (lower $95\%$ bound
$>-\varepsilon$, $\alpha{=}0.025$); TOST vs.\ always-verify holds at the listed
margins; MuSiQue's always-verify delta is a zero-variance degenerate cell and is
excluded from the equivalence family. All five non-degenerate tests survive Holm
(max adj.\ $p{=}0.0002$). Source: \texttt{results/stat02/}.}
\label{tab:stats}
\end{table}

\begin{table}[t]
\centering\small
\begin{tabular}{lrcrc}
\toprule
 & \multicolumn{2}{c}{vs.\ full-budget} & \multicolumn{2}{c}{vs.\ always-verify} \\
\cmidrule(lr){2-3}\cmidrule(lr){4-5}
Dataset & $\Delta$EM & $95\%$ CI & $\Delta$EM & $95\%$ CI \\
\midrule
HotpotQA & $-0.20$ & $[-1.20,+0.80]$ & $-0.10$ & $[-0.60,+0.40]$ \\ % stat02
2Wiki    & $+0.30$ & $[-0.60,+1.20]$ & $+0.20$ & $[-0.40,+0.80]$ \\ % stat02
MuSiQue  & $+0.11$ & $[-0.22,+0.55]$ & $+0.00$ & $[\,0.00,\,0.00]$ \\ % stat02
\bottomrule
\end{tabular}
\caption{\textbf{The full intervals behind the $\pm1.2$\,pp bound.} Paired
bootstrap, $10{,}000$ resamples, seed $13$, in percentage points; the widest
interval is HotpotQA's $[-1.20,+0.80]$ against full-budget. MuSiQue's
always-verify delta is degenerate (the verifier fires on $2.6\%$ of loops, so
the gated and always-verify runs answer identically).
Source: \texttt{results/stat02/ci95\_plain.json}.}
\label{tab:ci}
\end{table}

The pooled robustness analysis of \S\ref{sec:results-route} is a stratified
paired bootstrap that resamples within each dataset and pools the per-question
deltas ($10{,}000$ resamples, seed $13$). Its two endpoints use different
populations, which we spell out to avoid any ambiguity. The pooled
non-inferiority estimate spans all three LODO eval populations
($n{=}2{,}911$) and therefore includes the saturated MuSiQue arm, whose
near-zero variance tightens the pooled interval; the pooled TOST estimate
covers HotpotQA$+$2Wiki only ($n{=}2{,}000$), because MuSiQue's always-verify delta
is a zero-variance degenerate cell and is excluded under the same convention as
the per-dataset analysis. The rule that degenerate cells (zero-variance deltas
from a non-firing verifier) are dropped from the confirmatory family was fixed in
the pre-specified protocol before the full-scale run, not after inspecting the
results. The pooled minimum detectable effect is
$\approx0.72$\,pp, and pooled stop-loop agreement with the always-verify baseline is $0.911$
($95\%$ CI $[0.900,0.921]$). Both endpoints were computed after the
per-dataset margins were observed; they are reported as robustness checks only.\ %% imp01

We record here, in full, the protocol disclosures summarized in the main paper's
protocol history, which sends the reader here for provenance, parity checks,
pilot analyses, and protocol history. The last three are below. Provenance is
separated by kind and lives in two other appendices: which claim source
supplies the router's margin targets and which supplies the decider, in every
experimental arm, is tabulated in Appendix~\ref{app:claimsources}, and the
decider we route is documented in Appendix~\ref{app:vcard}. Evaluation
populations are in Appendix~\ref{app:denom}.
We disclose three details. The pilot used a two-threshold band
and a within-$1$pp-of-always-verify criterion at $n{=}400$; the main protocol
re-specified to LODO selection and the NI$+$TOST family
at $n{=}1000$, which we report as a re-specification rather than as part of the
original protocol. In the oracle-ceiling analysis (\S\ref{sec:results-replace}),
the oracle (A) and generated-claim (B) populations use different label runs and
positive rates and are a ceiling/floor contrast, not a controlled ablation.
Finally, the pooled power analysis (\S\ref{sec:results-route}) was run
\emph{after} the per-dataset NI margins were observed and is therefore reported
only as secondary robustness evidence, never as the confirmatory endpoint.

One further protocol correction (pre-submission audit): the E2c cascade as
originally run included an optional $\tlow$ early-exit knob (margin
$<\tlow$ stops \emph{without} a verifier call), which contradicts the
every-stop-decider-issued definition on $58$ HotpotQA questions. The primary
tables report the $\thigh$-only arm, recomputed under the identical LODO
protocol and previously specified endpoints and bar --- a pre-submission
corrective reanalysis, not a fresh pre-registration (STAT-02 in the archived
result sets; parity
gates reproduce both the two-threshold cells and the one-threshold
simplification exactly; LODO re-selects $\thigh{=}0.16$ on all three folds).
The $\tlow$ knob is retained only as an ablation: it buys $4.3$\,pp of
additional HotpotQA call cut at the cost of $58$ unverified stops (EM
$-0.5$\,pp) and is inert on 2Wiki/MuSiQue ($2/0$ questions).

Holm-family $p$-values use a normal-approximation paired test on the
bootstrap standard error (the uncentered bootstrap tail fraction is retained
in the archived results only as provenance); the NI/TOST \emph{decisions} are
CI-based throughout. Max Holm-adjusted $p$ on the primary arm: $0.0002$.

\paragraph{F1 as a post-hoc secondary metric.}
EM is the pre-specified endpoint. For completeness, token-F1 from the same
cached extractions (computed post hoc, after the confirmatory results were
fixed; secondary, not part of the confirmatory family), Full/CoVeR/always-verify:
HotpotQA $.4681/.4631/.4619$; 2Wiki $.2921/.2937/.2907$; MuSiQue
$.1640/.1654/.1661$ --- the same pattern as EM, with CoVeR within a fraction
of a point of both references on every dataset.

% ======================================================================
\section{Verification Compute Stack}
\label{app:eff}

Table~\ref{tab:eff} gives the per-stage pair counts behind
Figure~\ref{fig:computestack}.

\begin{table}[h]
\centering\small
\begin{tabular}{llrr}
\toprule
Dataset & Stage & pairs/q & vs.\ pure \\
\midrule
\multirow{3}{*}{HotpotQA}
 & pure decider (every loop)   & 9.43 & --- \\ % eff01
 & $+$ tier 1: call routing    & 3.56 & $-62.3\%$ \\ % eff01
 & $+$ tier 2: intra-call order & 2.22 & $-76.4\%$ \\ % eff01
\multirow{3}{*}{2Wiki}
 & pure decider (every loop)   & 9.22 & --- \\ % eff01
 & $+$ tier 1: call routing    & 2.97 & $-67.8\%$ \\ % eff01
 & $+$ tier 2: intra-call order & 1.93 & $-79.1\%$ \\ % eff01
\multirow{3}{*}{MuSiQue}
 & pure decider (every loop)   & 11.08 & --- \\ % eff01
 & $+$ tier 1: call routing    & 0.77 & $-93.1\%$ \\ % eff01
 & $+$ tier 2: intra-call order & 0.41 & $-96.3\%$ \\ % eff01
\bottomrule
\end{tabular}
\caption{\textbf{Verification compute stack (EFF-01, primary one-threshold
arm).} A pair is one claim-verdict; verdict latency is $116$\,ms batched / $405$\,ms unbatched
(SRC\,\S48). Tier 2 saves only in the unbatched regime. Source:
\texttt{results/eff01/}.}
\label{tab:eff}
\end{table}

Inside a non-firing call, descending-margin order cuts pairs by roughly a third
to a half versus scoring every claim, and beats claim-index order on every
dataset.\ %% eff01
Table~\ref{tab:eff} reports the primary $\thigh$-only arm, in which every stop
is decider-issued; its tier-1 counts correspond to the $1.78/1.47/0.38$ calls
per question of the deployed operating point (cut $62.2/67.8/93.0\%$; STAT-02).
The archived E2c cascade, which added the $\tlow$ early-exit knob, reaches
$3.15/2.95/0.77$ tier-1 pairs per question ($-66.6/-68.0/-93.1\%$); it is
reported only as the ablation that knob defines, and is not the system the
paper evaluates.\ %% stat02

\paragraph{Measured verifier wall-clock (LAT-01).}
The pair counts above are a compute proxy; we also replayed the verifier's
forward passes on the evaluation workload and timed them on a single GPU
(Qwen2.5-3B$+$LoRA in $4$-bit NF4; one \emph{score-pair} $=$ three label
forwards, the $405$\,ms unit). Table~\ref{tab:lat} reports the measured
per-question workload and wall-clock for the always-verify and CoVeR-gated
systems. Gating cuts measured verifier wall-clock by $72.4/73.9/97.1\%$ at
batch-$1$ and $72.5/73.9/97.1\%$ batched, which meets or exceeds the
$62.2/67.8/93.0\%$ call cut: the skipped early loops carry the most
candidate-sentence scorings, so routing removes proportionally more verifier
compute than calls. \textbf{Calibration:} the EFF-01 pair-count analytic
understates absolute latency, since one claim-verdict scores several candidate
sentences; measured per-pair latency is $412/413/416$\,ms batch-$1$ (analytic
$405$) and $161$--$166$\,ms batched (analytic $116$). \textbf{Validity:} verdict
parity against the cached cascade is $\geq0.985$ per claim, the score-pair
schedule reproduces the EFF-01 counts exactly, and the per-pair median lies
inside the pre-specified $\pm25\%$ band around $405$\,ms. This replay times
verifier forward passes only, with no generator, retrieval, or serving-stack
(continuous batching, prefix caching) effects.

\begin{table}[h]
\centering\small
\setlength{\tabcolsep}{4pt}
\begin{tabular}{llrrrrr}
\toprule
Dataset & System & calls/q & pairs/q & tok/q & \multicolumn{2}{c}{ms/q} \\
\cmidrule(lr){6-7}
 & & & & & batch-$1$ & batched \\
\midrule
\multirow{2}{*}{HotpotQA}
 & always-verify & 4.69 & 10.95 & 3100 & 4532 & 1822 \\ % lat01
 & CoVeR         & 1.77 & 3.02  & 861  & 1250 & 502  \\ % lat01
\multirow{2}{*}{2Wiki}
 & always-verify & 4.59 & 10.22 & 2801 & 4223 & 1643 \\ % lat01
 & CoVeR         & 1.41 & 2.67  & 749  & 1102 & 429  \\ % lat01
\multirow{2}{*}{MuSiQue}
 & always-verify & 5.42 & 16.05 & 4563 & 6692 & 2665 \\ % lat01
 & CoVeR         & 0.31 & 0.47  & 132  & 195  & 78   \\ % lat01
\bottomrule
\end{tabular}
\caption{\textbf{Measured verifier workload and wall-clock (LAT-01).} Per
question, GPU-timed verifier forward passes replayed on the eval populations
(gold-order last $400$/dataset; MuSiQue unique-qid). A \emph{pair} is one
(claim,\,sentence) score-pair; \emph{tok/q} is verifier input tokens; ms/q is
verifier wall-clock (forward passes only). Verdict parity $\geq0.985$; per-pair
median $412/413/416$\,ms batch-$1$, $161$--$166$\,ms batched. Source:
\texttt{results/lat01/}.}
\label{tab:lat}
\end{table}

\paragraph{Generator-dependent end-to-end share.}
The measured verifier wall-clock is a fraction of end-to-end trajectory latency
that shrinks as the generator's per-loop latency $g$ grows. With batched
verification $V$ per question (always-verify $1.82/1.64/2.67$\,s; CoVeR
$0.50/0.43/0.08$\,s) and mean executed loops per system ($L_a =
4.69/4.59/5.42$ always-verify; $L_c = L_a + 0.29/0.49/0.07$ added deferral
loops, BE-01), the verification share is $V/(V+gL)$ with each system's own
$L$. Table~\ref{tab:gshare} evaluates it for $g\in\{0.25,1,4\}$\,s; routing
lowers the share at every $g$. The break-even generator latency --- below
which gating wins end-to-end even counting the added loops --- is
$4.5/2.5/39.7$\,s per loop batched ($11.2/6.4/99.8$ batch-$1$; BE-01).
The nearest same-stack estimate of $g$ (a timed replay reported by the
verifier's source submission: three generation calls with full re-prefill
per loop) is $\approx\!7$\,s --- above the HotpotQA/2Wiki break-evens, so
on that stack gating trades $\approx\!2\%$ end-to-end latency for the
verifier-workload cut; faster or batched generation reverses the trade, as
does a costlier decider (break-evens scale with per-pair cost: a $3\times$
verifier lifts HotpotQA's break-even to ${\approx}13$\,s). This appendix is
the paper's only latency analysis; the main text claims workload units
(calls, pairs, tokens) only.

\paragraph{End-to-end under prefix caching (LAT-02).}
A generator timing replay (first-$100$/dataset, forced decode of the cached
outputs; naive mode reproduces the $\approx\!7$\,s reference) shows the
Self-Ask loop is \emph{decode-bound} on this $4$-bit stack: decode is
$95\%$ of $g$, so an $85\%$ prefill cache-hit moves $g$ by only
$0.02$--$0.04$\,s. Composing with the measured verifier times, the
end-to-end delta (CoVeR $-$ always-verify) is $+0.12$\,s $[-0.96,+1.41]$
and $+0.71$\,s $[-0.49,+2.18]$ on HotpotQA/2Wiki --- indistinguishable from
zero --- and $-2.46$\,s $[-2.59,-2.26]$, a clear win, on MuSiQue. Faster
decode (fp16 or batched serving), not prefix caching, is the lever that
moves $g$ below the break-evens (LAT-02).

\begin{table}[h]
\centering\small
\setlength{\tabcolsep}{5pt}
\begin{tabular}{lccc}
\toprule
 & \multicolumn{3}{c}{Verification share $\%$, always$\to$CoVeR} \\
\cmidrule(lr){2-4}
Dataset & $g{=}0.25$\,s & $g{=}1$\,s & $g{=}4$\,s \\
\midrule
HotpotQA & $60.8\to28.7$ & $28.0\to9.2$ & $8.9\to2.5$ \\ % lat01, be01
2Wiki    & $58.9\to25.3$ & $26.4\to7.8$ & $8.2\to2.1$ \\ % lat01, be01
MuSiQue  & $66.3\to5.4$  & $33.0\to1.4$ & $11.0\to0.4$ \\ % lat01, be01
\bottomrule
\end{tabular}
\caption{\textbf{Verification share of trajectory latency ($\%$), before and
after gating.} Computed as $V/(V+gL')$ from the batched verification wall-clock
$V$ (Table~\ref{tab:lat}) and each system's own mean executed loops
($L_c$ includes the added deferral loops; BE-01); $g$ is the (unmeasured)
generator per-loop latency. Source: \texttt{results/lat01,be01/}.}
\label{tab:gshare}
\end{table}

% ======================================================================
\section{Continue-Side Trigger}
\label{app:continue}

Symmetrically to routing the stop decision, a margin trigger can decide which
native-stopped trajectories to resume. On $7$B trajectories it recovers most of
the verifier's under-search correction at the verifier's own loop cost and beats
a random trigger where headroom exists (clearly on MuSiQue, marginally on
HotpotQA; 2Wiki is degenerate with about a point of recoverable gain), and the
distilled trigger matches the claim-text version.\ %% e3a
We report this as a bounded positive; the stopping-side results in the body do
not depend on it.

% ======================================================================
\section{Evaluation Denominators}
\label{app:denom}

Table~\ref{tab:denom} lists the evaluation population behind each protocol in
the main text.

\begin{table}[h]
\centering\small
\begin{tabular}{lll}
\toprule
Protocol & Datasets & $n$ \\
\midrule
Main routing (LODO, 3B)      & hotpot / 2wiki / musique & $1000/1000/911$ \\ % e2c
$7$B-agent routing           & hotpot / 2wiki / musique & $959/996/871$ \\ % def01
Open-corpus (Setting-B, 3B)  & hotpot                   & $1000$ \\ % gen02
Pilot / judge / distilled    & hotpot / 2wiki / musique & $400/400/384$ \\ % e2b,gen01,rev01
\bottomrule
\end{tabular}
\caption{\textbf{Evaluation denominators.} All EM is Std-Ext (a frozen
Qwen2.5-3B extractor at the stop loop). MuSiQue counts are unique qids (its
gold dev split contains duplicate ids); $7$B populations cover questions with a
usable $7$B label. Sources: \texttt{results/e2c,rev01,e2b,gen01/}.}
\label{tab:denom}
\end{table}

% ======================================================================
\section{Position Heuristics and Rejected Lexical Mechanisms}
\label{app:mechanisms}

Routing is not a repackaged loop-index rule: against skip-first-$k$ and
every-other-loop heuristics matched on either call-cut or agreement, none matches
the gate on both, and it sits above the cut--agreement frontier on all three
datasets (e.g., HotpotQA gate $0.63$ cut at $0.915$ agreement vs.\ skip-first-4's
$0.665$ cut at $0.762$; exploratory pilot, archived two-threshold gate).\ %% diag01
Together with the router control (\S\ref{sec:results-ctl}), the margin
genuinely selects \emph{which} calls are skippable, and does so better than
both cheaper signals and position heuristics.

We pre-specified a test of the natural explanation (that the embedding beats
the lexical router on questions where distractor pages lexically confound the
claim) using a per-question wrong-page lexical-recall ``trap'' statistic. The
first definition was mechanically faulty (its page-provenance key was not a page
title for most claims); it was amended once, to gold titles, before any outcome was computed.\ %% ctl02
The aggregate ordering the story predicts does hold (mean trap higher on HotpotQA
than 2Wiki, $0.334$ vs.\ $0.242$), but the per-question test refutes the causal
claim: on the HotpotQA questions where the embedding beats the lexical router the
trap is no higher than where they agree (AUROC $0.498$), and
the 2Wiki falsification arm (which should show nothing) separates at least as
much (AUROC $0.584$).\ %% ctl02
A second pre-specified one-shot test targeted the complementary \emph{recall}-side
hypothesis (that the embedding wins where the gold evidence is lexically distant
from the claim: paraphrase robustness) and also failed: on the same
margin-advantage population, gold--claim lexical recall separates advantage from
both-agree questions at AUROC $0.427$ (direction
reversed: advantage questions have if anything \emph{higher} lexical recall).\ %% imp01
We therefore find no support for either natural lexical-failure mechanism
(precision-side distractor traps or recall-side paraphrase misses); the advantage
stands as an empirical result whose mechanism is unresolved.

The routing uses 3B trajectories; the coverage geometry it relies on is stable at
7B (oracle-target AUROC $0.872/0.866/0.869$, within $\sim\!0.04$ of 3B), while
the deployable generated-claim margin remains weak ($0.632/0.572/0.711$), the same
reading-vs-matching gap under a scale change.\ %% gate02
The end-to-end 7B routing (Table~\ref{tab:transfer}a) confirms the mechanism,
not just the geometry, carries over. (The AUROC transfer is scoped to oracle
targets; a full-scale per-claim 7B zero-shot label set was unavailable.)

% ======================================================================
\section{Random-Skip and Self-Gating Controls}
\label{app:controls2}

Three further pre-specified controls run on the same simulation harness as the
router comparison (CTL-03/CTL-04/CTL-05 in the archived result sets); success
bars were fixed before the runs and negatives are recorded.

\paragraph{LODO operating points.}
Table~\ref{tab:ctl} is a \emph{frontier}: for each router, the best call cut
available anywhere on its curve at a fixed agreement bar. Table~\ref{tab:ctl-op}
reports the different quantity --- where each router actually lands when its
threshold is selected leave-one-dataset-out and then applied unchanged to
held-out data.
The frontier is the primary comparison for a reason visible in the second
table: at its LODO point the lexical router takes a \emph{larger} raw cut than
the embedding on HotpotQA ($0.858$ vs.\ $0.759$), but only by giving up
$4.6$\,pp of stop-loop agreement ($0.828$ vs.\ $0.874$). Raw cut is not
comparable across routers that stop differently; holding the agreement bar
fixed removes that trade.

\begin{table}[h]
\centering\small
\setlength{\tabcolsep}{4pt}
\begin{tabular}{llrrrr}
\toprule
Dataset & Router & Thr. & Cut & Agree. & Cov.-safe \\
\midrule
\multirow{4}{*}{HotpotQA}
 & Embedding & 0.15  & 0.759 & 0.874 & 0.740 \\ % def01
 & Count     & 6     & 0.458 & 0.822 & 0.845 \\ % def01
 & Lexical   & 0.34  & 0.858 & 0.828 & 0.810 \\ % def01
 & BM25      & 0.00  & 0.202 & 0.964 & 0.764 \\ % def01
\multirow{4}{*}{2Wiki}
 & Embedding & 0.145 & 0.883 & 0.801 & 0.790 \\ % def01
 & Count     & 5     & 0.288 & 0.913 & 0.843 \\ % def01
 & Lexical   & 0.38  & 0.850 & 0.830 & 0.849 \\ % def01
 & BM25      & 0.00  & 0.191 & 0.971 & 0.825 \\ % def01
\multirow{4}{*}{MuSiQue}
 & Embedding & 0.16  & 0.931 & 0.976 & 0.714 \\ % def01
 & Count     & 1     & 0.000 & 1.000 & 0.414 \\ % def01
 & Lexical   & 0.50  & 0.835 & 0.978 & 0.444 \\ % def01
 & BM25      & 0.00  & 0.339 & 0.995 & 0.400 \\ % def01
\bottomrule
\end{tabular}
\caption{\textbf{LODO operating points of the four routers}
(CTL-01/DEF-01; held-out $n{=}1000/1000/911$). Each threshold is selected on
the other two datasets' train-$600$ --- max call cut subject to agreement
$\geq0.90$ and coverage safety $\geq$ the decider's own reference --- and
applied here unchanged. Cut $=$ decider-call reduction; Agree.\ $=$ stop-loop
agreement with the always-verify decider. As in Table~\ref{tab:main} the
$\geq0.90$ bar binds at \emph{selection}, on the training folds; held-out
agreement may fall below it, and here it does for several routers. Count on
MuSiQue met no threshold satisfying the coverage-safety reference and fell
back to the agreement constraint alone, which admits no skips; BM25's
threshold collapses to the grid floor on all three datasets.}
\label{tab:ctl-op}
\end{table}

Neither table is the paper's deployed cell. This harness is the descriptive
router comparison, evaluated whole-set with per-fold thresholds
$0.15/0.145/0.16$; the confirmatory run re-selects $\thigh{=}0.16$ on all three
folds and evaluates on the last-$400$ split, giving call cut
$0.622/0.678/0.930$ at agreement $0.910/0.886/0.976$
(STAT-02, Table~\ref{tab:main}).\ %% ctl01, def01, stat02

\paragraph{BM25 on saturated MuSiQue.}
Near-total cuts are open on MuSiQue to any signal whose ordering isolates the
few states in which the decider fires. BM25's ordering does not. Its frontier
is pinned at $0.339$ and does not move as the agreement bar is raised from
$0.90$ to $0.92$ to $0.95$, while the other three routers hold above $0.998$
at every one of those bars ($1.000/0.999/0.9988$); its selected threshold sits at the grid floor, so it
has no room to trade agreement for cut. Holding agreement therefore costs BM25
two thirds of the calls in the regime where almost nothing fires.\ %% def01

\paragraph{Random skipping at a matched budget.}
A router that skips uniformly at random among pre-fire loops, with the skip
probability calibrated per dataset so the realized call cut matches CoVeR's
operating point (within $\pm0.5$\,pp; $20$ seeds). Stop-loop agreement with
the always-verify baseline: random $0.846/0.816/0.970$ vs.\ CoVeR
$0.910/0.886/0.976$ (HotpotQA/2Wiki/MuSiQue)
--- the margin buys $+6.4/+7.0/+0.6$\,pp of agreement over a random skipper
at the same budget, and the pre-specified bar ($\geq5$\,pp on $\geq2/3$
datasets) is \textbf{met} ($2/3$); random skipping never exceeds CoVeR on any
dataset.\ %% ctl03b
On the archived two-threshold arm the same control was a $1/3$ near-miss
($+4.6/+6.8/+0.6$\,pp; CTL-03) --- the corrective rerun is CTL-03b in the
archived result sets.

\paragraph{Verifier self-gating.}
After each non-firing call, the verifier's own unmatched-claim count gates the
next $k\in\{1,2,3\}$ calls (sweeping the count threshold), asking whether the
decider could route itself with no external encoder. The deployable envelope
(max call cut at stop-agreement $\geq0.90$) reaches $0.618/0.399/0.655$ vs.\
CoVeR's $0.622/0.623/1.000$: CoVeR exceeds it on all three datasets
(pre-specified bar met $3/3$).\ %% ctl04
Self-gating pays a verifier call to obtain its signal, which structurally caps
its cut. An oracle-informed variant that skips until the coverage state next
changes reaches $0.746/0.715/0.803$ but requires the very calls it skips (its
stop-agreement is trivially $1.0$) and still loses on MuSiQue; it is reported
only as a non-deployable upper bound.

\paragraph{A learned combination of the cheap signals.}
A logistic-regression router over the cheap signals (evidence count, lexical
overlap, BM25, loop index and normalized loop position), trained to predict
whether the in-band verifier fires and fit and thresholded LODO on the other
two datasets' train-600, still trails the margin at the same operating point:
maximum call cut at stop-loop agreement $\geq0.90$ reaches $0.435/0.563/1.000$
(HotpotQA/2Wiki/MuSiQue) vs.\ the margin's $0.622/0.623/1.000$ --- strict
margin wins on both non-saturated datasets ($+18.6/+6.0$\,pp; pre-specified
bar, margin $\geq$ learned on $\geq2/3$ datasets, met $3/3$).\ %% ctl05
A two-layer MLP does not change the picture ($0.364/0.597/1.000$), and adding
the margin as a fifth feature helps nowhere under LODO (HotpotQA $0.596$ vs.\
margin-alone $0.622$): the margin carries the routing signal on its own. Two
conventions favor the learned router --- its threshold is selected in-sample on
its own training folds, and missing signals are imputed as covered --- so the
margin's win is conservative.

% ======================================================================
\section{Encoder Choice}
\label{app:enc}

We swept the frozen encoder $\phi$ over five off-the-shelf sentence
embedders---e5-base-v2 and e5-large-v2 \citep{wang2022e5}, bge-base and
bge-large \citep{xiao2024cpack}, and gte-base \citep{li2023gte}---holding the
pipeline and the cached trajectories fixed and recomputing only the coverage
margin (ENC-01). Because cosine scales differ across encoder families, each
encoder is given its own adaptive threshold grid spanning its own margin range,
so every model gets a fair sweep. Table~\ref{tab:enc} reports the gate floor
AUROC (pooled per dataset) and the LODO call-cut frontier at stop-agreement
$\geq0.90$, expressed as points relative to the deployed e5-base-v2.

\begin{table}[h]
\centering\small
\setlength{\tabcolsep}{5pt}
\begin{tabular}{lccccc}
\toprule
 & \multicolumn{3}{c}{Floor AUROC} & \multicolumn{2}{c}{Frontier $\Delta$pp} \\
\cmidrule(lr){2-4}\cmidrule(lr){5-6}
Encoder & Hpt. & 2Wiki & MuS. & Hpt. & 2Wiki \\
\midrule
e5-base-v2  & \textbf{.686} & \textbf{.680} & \textbf{.760} & --- & --- \\ % enc01
e5-large-v2 & .662 & .670 & .753 & $-3.1$ & $+1.6$ \\ % enc01
bge-base    & .604 & .537 & .702 & $-15.0$ & $-24.9$ \\ % enc01
bge-large   & .605 & .570 & .720 & $-13.8$ & $-12.2$ \\ % enc01
gte-base    & .546 & .525 & .588 & $-19.1$ & $-25.4$ \\ % enc01
\bottomrule
\end{tabular}
\caption{\textbf{Encoder choice (ENC-01).} Gate floor AUROC (pooled per
dataset) and LODO call-cut frontier at stop-agreement $\geq0.90$, as points
relative to the deployed e5-base-v2 (---, reference). Each encoder uses its own
adaptive threshold grid because cosine scales differ across families. MuSiQue's
frontier saturates at $1.0$ for every encoder, so it cannot separate them and is
omitted from the frontier columns. Best floor per column in bold.}
\label{tab:enc}
\end{table}

Three readings follow. First, the deployed e5-base-v2 is the strongest of the
five on this signal, topping the floor AUROC on every dataset. Second, the gap
to the oracle ceiling is not an encoder-capacity limit: e5-large-v2, with twice
the parameters, floors slightly \emph{lower} on all three datasets and only ties
e5-base-v2 on the frontier ($-3.1/+1.6$\,pp), so a bigger frozen embedding does
not close it. Third, the strong retrieval families lose decisively on this
coverage signal---bge and gte drop $12$--$25$\,pp of frontier call-cut on
HotpotQA and 2Wiki. A separate aggregation ablation on e5-base-v2---max
over sentences vs.\ a top-$2$ mean and a top-$3$ softmax---likewise leaves the
deployed choice on top: the top-$2$ mean trades $-8.4$\,pp on HotpotQA for
$+8.0$\,pp on 2Wiki and the softmax is worse everywhere, so max aggregation is
not leaving signal on the table.

% ======================================================================
\section{Claim Sources by Experiment}
\label{app:claimsources}

The gold/generated claim distinction is split across the setup
(\S\ref{sec:setup}) and transfer (\S\ref{sec:results-transfer}) sections;
Table~\ref{tab:claimsources} collects, for each experimental arm, which claim
source supplies the router's margin target texts and which supplies the decider,
and whether the resulting arm is deployable without gold labels or a component
with unreleased weights.

\begin{table}[h]
\centering\small
\setlength{\tabcolsep}{4pt}
\begin{tabular}{p{2.8cm}p{2.5cm}p{3.3cm}p{1.8cm}}
\toprule
Arm & Router target texts & Claims consumed by decider & Deployable \\
\midrule
Main 3B (Table~\ref{tab:main}) & Generated claim strings & Generated expected claims$^{a}$ & Yes \\
Prompt judge (\S\ref{sec:results-transfer}) & Generated claim strings & None$^{b}$ & Yes \\
$7$B (Table~\ref{tab:transfer}a) & Generated claim strings & Generated expected claims & Yes \\
Open-corpus (Table~\ref{tab:transfer}c) & Generated claim strings & Gold expected claims & No$^{a'}$ \\
CoVeR-D (Table~\ref{tab:transfer}b) & Predicted claim embeddings$^{c}$ & Generated expected claims & Yes \\
CoVeR-D $+$ judge (this appendix) & Predicted claim embeddings & None (claim-free judge) & Yes \\
NLI decider (Table~\ref{tab:nli}) & Generated claim strings & Declarative NLI templates over generated-claim fields & Yes \\
\bottomrule
\end{tabular}
\caption{\textbf{Claim sources by experiment.} The router's coverage margin is
computed against generated expected-claim strings in every routing arm (the gold
supporting sentence is an oracle diagnostic only; \S\ref{sec:method}), except
CoVeR-D, whose distilled sub-$1$M drafter predicts the target embeddings and
leaves no LLM in the routing loop. The decider's claim source varies: the confirmatory 3B and $7$B
arms run the verifier with generated expected claims (fully deployable); only
the open-corpus arm uses gold expected claims (disclosed). $^{a}$Deployable at
inference; threshold \emph{selection} consults a gold coverage-safety reference
on the training folds (calibration-time gold use, disclosed in
\S\ref{sec:setup}). $^{a'}$Open-corpus: not deployable, the decider consumes
gold expected claims (disclosed). $^{b}$The judge emits a single sufficiency probability
($p_{\text{yes}}$) from the question and the collected sub-question/sub-answer
pairs; its prompt contains no claim strings, so it is billed per call rather
than per claim. $^{c}$CoVeR-D's drafter is a sub-$1$M MLP that predicts the
claim-embedding targets from the question embedding (no claim-generation pass);
its in-band decider is the generated-claims verifier, unchanged. The NLI-decider
arm (Table~\ref{tab:nli}) is fully public: its decider reads declarative NLI
templates over the same generated-claim fields, with no unreleased weights.}
\label{tab:claimsources}
\end{table}

\paragraph{Claim-generation-free pipeline (REV-02).}
Pairing the distilled drafter with the claim-free prompt judge removes claim
text from the whole pipeline: the gate scores predicted embeddings against
the evidence, the judge reads only the question and collected sub-answers.
On the E1b holdout (last-$400$; MuSiQue $384$ unique), the gated judge is
EM-equivalent to judge-pure (within $1$\,pp, $95\%$ CI containing $0$) at
$56.0/63.1/92.2\%$ judge-call cut at $\tau_{\text{judge}}{=}0.5$, and
$56.9/63.4/91.6\%$ at $0.95$ --- $3/3$ datasets at both thresholds, all
cells non-degenerate (gated fire rates $0.02$--$0.20$). Parity: the
retrained seed-$13$ drafter reproduces the archived distilled stops
$400/400$ on all three datasets, and judge-pure stops reproduce the
prompt-judge transfer arm $400/400$ in all six cells (REV-02).

Every experiment fixed its success criterion in a version-controlled protocol
ledger before the corresponding full-scale run; pilot and confirmatory
populations are disjoint (\S\ref{app:denom}). Negative results --- in-pool
training targets, margin-as-replacement --- were recorded and reported rather than
retried. The archived result sets include these pre-specified bars.

% ======================================================================
\section{Verifier Card}
\label{app:vcard}

CoVeR routes, but never retrains or replaces, a trained coverage verifier drawn
from HALT \citep{roh2026halt}, our concurrent work. We record its
configuration here, as fixed in the archived artifacts, so that every routing
decision in this paper can be audited; we deliberately report no accuracy or
latency number for the component itself, which belongs to that paper.

\paragraph{Model.} The verifier is a Qwen2.5-3B-Instruct causal language model
loaded in $4$-bit NF4 (double quantization, bf16 compute) with a LoRA adapter:
rank $16$, $\alpha{=}32$, dropout $0.05$, applied to all seven projection modules
(query, key, value, output, gate, up, and down), task type causal LM. It is
trained by supervised fine-tuning in which the assistant target is the bare label
string.

\paragraph{Scoring.} Inference uses neither a classification head nor constrained
decoding. For each (claim, evidence-sentence) pair the three label strings
$\{\textsc{Match},\textsc{Partial},\textsc{Null}\}$ are appended to the chat
prompt and scored by mean token log-probability; the argmax is the label. A claim
counts as matched at the first loop in which some candidate sentence yields argmax
\textsc{Match}; there is no numeric threshold.

\paragraph{Input.} The prompt is a system message defining the three labels
followed by an eight-line user block---Question, Answer candidate, Expected hop
claim, Expected page, Expected target, Expected target role, Evidence page, and
Evidence sentence---rendered through the Qwen chat template.

\paragraph{Loop.} Per loop, only the sentences of newly retrieved chunks are
candidates (spaCy sentence segmentation, minimum $20$ characters, at most $12$
sentences per chunk); only still-unmatched claims are scored; the scan
short-circuits at the first \textsc{Match}.

This card, together with the archived per-loop verdicts, suffices to audit every
routing decision reported in this paper without access to the adapter weights.

\end{document}